\documentclass[10pt,twocolumn,letterpaper]{article}

\usepackage{graphicx}
\usepackage{amsmath}
\usepackage{subfigure}
\usepackage{fancyhdr}
                       
\graphicspath{{.}}
\DeclareGraphicsExtensions{.pdf}

\newcommand{\ie}{{{\em i.e.} }}
\newcommand{\eg}{{{\em e.g.} }}
\newcommand{\vs}{{{\em vs} }}
\newcommand{\fig}[1]{{Fig. \ref{fig:#1}}}
\newcommand{\tab}[1]{{Tab. \ref{tab:#1}}}
\newcommand{\sect}[1]{{Sect. \ref{sec:#1}}}
\newcommand{\myindex}[1]{{#1}}

\begin{document}

\title{From Virtual to Real World Visual Perception using Domain Adaptation -- The DPM as Example}

\author{Antonio M. L\'{o}pez\\
Computer Vision Center (CVC) and Dpt. Ci\`{e}ncies de la Computaci\'{o} (DCC),\\ 
Universitat Aut\`{o}noma de Barcelona (UAB)\\
{\tt\small antonio@cvc.uab.es}
\and
Jiaolong Xu\\
CVC and DCC, UAB\\
{\tt\small jiaolong@cvc.uab.es}
\and
Jos\'{e} L. G\'{o}mez\\
CVC and DCC, UAB\\
{\tt\small jlgomez@cvc.uab.es}
\and
David V\'{a}zquez\\
CVC and DCC, UAB\\
{\tt\small dvazquez@cvc.uab.es}
\and
Germ\'{a}n Ros\\
CVC and DCC, UAB\\
{\tt\small gros@cvc.uab.es}
}

\maketitle

\abstract{Supervised learning tends to produce more accurate classifiers than unsupervised learning in general. This implies that training data is preferred with annotations. When addressing visual perception challenges, such as localizing certain object classes within an image, the learning of the involved classifiers turns out to be a practical bottleneck. The reason is that, at least, we have to frame object examples with bounding boxes in thousands of images. A priori, the more complex the model is regarding its number of parameters, the more annotated examples are required. This annotation task is performed by human oracles, which ends up in inaccuracies and errors in the annotations ({\em aka} ground truth) since the task is inherently very cumbersome and sometimes ambiguous. As an alternative we have pioneered the use of {\em virtual worlds} for collecting such annotations automatically and with high precision. However, since the models learned with virtual data must operate in the real world, we still need to perform {\em domain adaptation} (DA). In this chapter we revisit the DA of a {\em deformable part-based model} (DPM) as an exemplifying case of virtual- to real-world DA. As a use case, we address the challenge of {\em vehicle detection} for driver assistance, using different publicly available virtual-world data. While doing so, we investigate questions such as: how does the domain gap behave due to virtual-vs-real data with respect to dominant object appearance per domain, as well as the role of photo-realism in the virtual world.}

\section{Need for Virtual Worlds}
\label{sec:needVW-virtualDPM}

Since the 90's, {\em machine learning} has been an essential tool for solving {\em computer vision} tasks such as image classification, object detection, instance recognition, and (pixel-wise) semantic segmentation, among others \cite{ViolaCVPR01Rapid,DalalCVPR05Histograms,FelzenszwalbPAMI10Object,SanchezIJCV13Image,BengioPAMI13Representation}. In general terms, the best performing machine learning algorithms for these tasks are {\em supervised}; in other words, not only the raw data is required, but also {\em annotated information}, \ie {\em ground truth}, must be provided to run the training protocol. Collecting the annotations has been based on human oracles and collaborative software tools such as Amazon's Mechanical Turk \cite{AmazonMT}, LabelMe \cite{RussellIJCV08LabelMe}, etc. It is known, that human-based annotation is a cumbersome task, with ambiguities, and inaccuracies. Moreover, not all kinds of ground truth can be actually collected by relying on human annotators, \eg pixel-wise optical flow and depth. 

\begin{figure*}
\includegraphics[width=\textwidth]{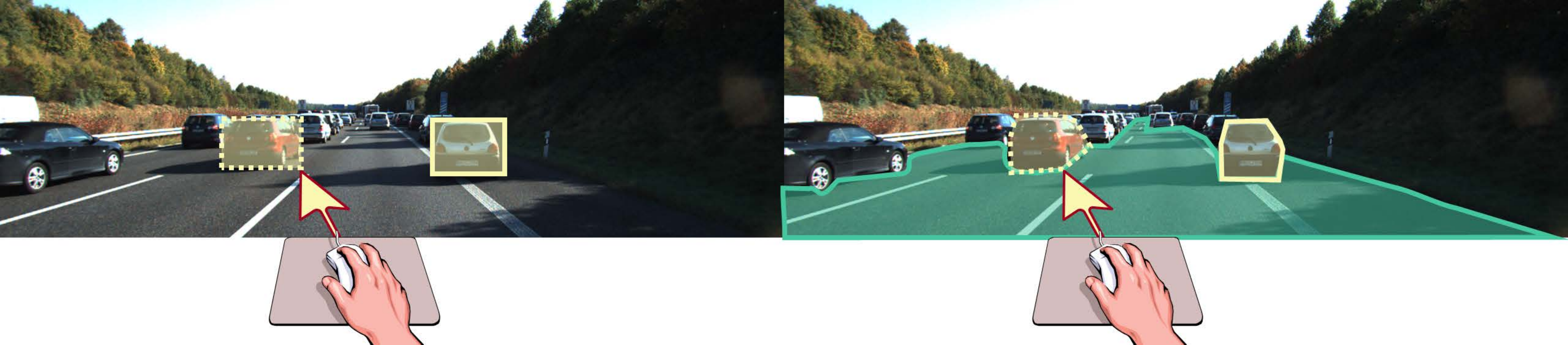}
\caption{Ground truth obtained by human annotation: {\em left)} framing the rectangular bounding box (BB) of vehicle instances; {\em right)} delineating the  contours (silhouettes) between the different classes of interest contained in the image, even at instance level.}
\label{fig:HumanAnnotation}
\end{figure*}

The non-expert reader can have a feeling of the annotation effort by looking at \fig{HumanAnnotation}, where we can see two typical annotation tasks, namely bounding box (BB) based object annotations, and delineation of semantic contours between classes of interest. In the former case, the aim is to develop an object detector (\eg a vehicle detector); in the latter, the aim is to develop a pixel-wise multi-class classifier, \ie to perform the so-called semantic segmentation of the image. 

With the new century, different datasets were created with ground truth and put publicly available for research. Providing a comprehensive list of them is out of the scope of this chapter, but we can cite some meaningful and pioneering examples related to two particular tasks in which we worked actively, namely {\em pedestrian detection} and {\em semantic segmentation}; both in road scenarios for either {\em \myindex{advanced driver assistance systems}} (ADAS) or {\em \myindex{autonomous driving}} (AD). One example is the Daimler Pedestrian dataset \cite{EnzweilerPAMI09Monocular}, which includeds 3,915 BB-annotated pedestrians and 6,744 pedestrian-free images (\ie image-level annotations) for training, and 21,790 images with 56,492 BB-annotated pedestrians for testing. Another example corresponds to the pixel-wise class ground truth provided in \cite{BrostowPRL09Semantic} for urban scenarios; giving rise to the well-known {\em CamVid} dataset which considers 32 semantic classes (although only 11 are usually considered) and includes 701 annotated images, 300 normally used for training and 401 for testing. A few years after, the KITTI Vision Benchmark Suite \cite{GeigerIJRR13Vision} was an enormous contribution for the research focused on ADAS/AD given the high variability of the provided synchronized data (stereo images, LIDAR, GPS) and ground truth (object bounding boxes, tracks, pixel-wise class, odometry).

In parallel to these annotation efforts and the corresponding development of new algorithms (\ie new human-designed features, machine learning pipelines, image search schemes, etc.) for solving computer vision tasks, {\em deep leaning} was finding its way to become the powerful tool that is today for solving such tasks. Many researchers would point out \cite{KrizhevskyNIPS12Imagenet} as a main breakthrough, since deep {\em convolutional neural networks} (CNNs) showed an astonishing performance in the data used for the {\em ImageNet Large-Scale Visual Recognition Challenge} (ILSVRC). ImageNet \cite{DengCVPR09Imagenet} contains over 15 million of human-labeled (using Mechanical Turk) high-resolution images of roughly 22,000 categories. Thus, ImageNet was a gigantic human annotation effort. ILSVRC uses a subset of ImageNet with about 1,000 images of 1,000 categories; overall, about 1.2M images for training, 50,000 for validation, and 150,000 for testing. Many deep CNNs developed today rely on an ImageNet pre-trained deep CNN which is modified or fine-tuned to solve a new task or operate in a new domain. The research community agrees in the fact that, in addition to powerful GPU hardware to train and test deep CNNs, having a large dataset with ground truth such as ImageNet is key for their success. In this line, more recently, it was released MS COCO dataset \cite{LinECCV14Microsoft}, where per-instance object segmentation is provided for 91 object types on 328,000 images, for a total of 2.5M of labeled instances.

As a matter of fact, in the field of ADAS/AD we would like to have datasets with at least the variety of information sources of KITTI and the ground truth size of ImageNet/COCO. However, when looking at the ground truth of KITTI in quantitative terms, we can see that individually they are in the same order of magnitude than other ADAS/AD-oriented publicly available datasets (\eg see the number of pedestrians BBs of KITTI and Daimler datasets, and the number of pixel-wise annotated images of KITTI and CamVid). A proof of this need is the recently released Cityscapes dataset \cite{CordtsCVPR16Cityscapes} which tries to go beyond KITTI in several aspects. For instance, it includes 5,000 pixel-wise annotated (stereo) images covering 30 classes and per-instance distinction, with GPS, odometry and ambient temperature as metadata. In addition, it includes 20,000 more images but where the annotations are coarser regarding the delineation of the instance/class contours. This kind of dataset is difficult to collect since driving through 50 cities covering several months and weather conditions was required. Moreover, providing such a ground truth can take from 30 to 90 minutes per image for a human oracle in case of fine-grained annotations and depending on the image content. 

For the semantic segmentation task, Cityscapes goes one order of magnitude beyond KITTI and CamVid. However, it is far from the annotation numbers of ImageNet and MS COCO. The main reason is two-fold. On the one hand, data collection itself, \ie Cityscapes images are collected from on-board systems designed for ADAS/AD not just downloaded from an internet source; moreover, metadata such as GPS and vehicle odometry is important, not to mention the possibility of obtaining depth from stereo. On the other hand, the annotations must be more precise since ultimately the main focus of ADAS/AD is on reducing traffic accidents. In any case, as we mentioned before, other interesting ground truth types are not possible or really difficult to obtain by human annotation, \eg pixel-wise optical flow and depth (without active sensors); but eventually these are important cues for ADAS/AD based on visual perception.

In this ADAS/AD context, and due to the difficulties and relevance of having large amounts of data with ground truth for training, debugging and testing, roughly since 2008 we started to explore a different approach. In particular, the idea of using realistic virtual worlds (\eg based on videogames) for training vision-based perception modules. The advantages were clear: (1) forcing the driving and data acquisition situations needed; (2) obtaining different types of pixel-wise ground truth (class ID, instance ID, depth, optical flow); (3) generating such data relatively fast (\eg currently our \myindex{SYNTHIA} environment \cite{RosCVPR16Synthia} can generate 10,000 images per hour with such ground truths, using standard consumer hardware); etc. Of course, such a proposal also came with doubts such as {\em can a visual model learned in virtual worlds operate well in real-world environments?}, and {\em does this depend on the degree of photo-realism?}. From our pioneering paper \cite{MarinCVPR10Learning}, where we used pedestrian detection based on HOG/Linear-SVM as proof-of-concept, to our last work, \ie SYNTHIA \cite{RosCVPR16Synthia}, where we have addressed pixel-wise semantic segmentation via deep CNNs, we have been continuously exploring the idea of learning in virtual worlds to operate in real environments.

The use of synthetic data has attracted the attention of other researchers too, and more recently specially due to the massive adoption of deep CNNs to perform computer vision tasks and their data hungry nature. 3D CAD models have been used to train visual models for pose estimation, object detection and recognition, and indoor scene understanding \cite{GraumanICCV03Inferring, AgarwalACCV06Local, SatkinBMVC10Back, ShottonCVPR11Realtime, PishchulinCVPR11Learning, SchelsICMR11Synthetically, PepikCVPR12Teaching, SatkinBMVC12Data, AubryCVPR14Seeing, ChenCVPR14Beat, SunBMVC14Virtual, PaponICCV15Semantic, PanaredaBustoBMVC15Adaptation, PengICCV15Learning, HandaX15SynthCam3D, HattoriCVPR15Learning, RozantsevCVIU15Rendering, MassaCVPR16Deep, SuICCV163D, SuICCV16Render, MovshovitzX16Useful, BochinskiAVSS16Training, HandaCVPR16Understanding}; a  virtual racing circuit has been used for generating different types of pixel-wise ground truth (depth, optical flow and class ID) \cite{HaltakovGCPR13Framework}; videogames have been used for training deep CNNs with the purpose of semantic segmentation \cite{RichterECCV16Playing} and depth estimation from RGB \cite{ShafaeiBMVC16Play}; synthetic scenarios have been used also for evaluating the performance of different feature descriptors \cite{KanevaICCV11Evaluating, AubryICCV15Understanding, VeeravasarapuX15Simulations, VeeravasarapuX15Model, VeeravasarapuX16Model} and for training and testing optical and/or scene flow computation methods \cite{MeisterCEMT11Real, ButlerECCV12Naturalistic, OnkarappaMTA15Synthetic, MayerCVPR16Large}, stereo algorithms \cite{HaeuslerGCPR13Synthesizing}, as well as trackers \cite{TaylorCVPR07OVVV}, even using synthetic clones of real-world areas of interest \cite{GaidonCVPR16Virtual}; synthetic LIDAR-style data has been used for object detection too \cite{LaiRSS09Laser, LaiIJRR10Object}; finally, virtual worlds are being used for learning high-level artificial behavior such as playing Atari games \cite{MnihNIPSWDL13Playing}, reproducing human behavior playing shooter games \cite{LlarguesESA14Artificial} and driving/navigating end-to-end \cite{ChenICCV15DeepDriving, ZhuX16Target}, even learning {\em unwritten} common sense \cite{VedantamICCV15Learning, ZitnickPAMI16Adopting}.

\section{Need for Domain Adaptation}
\label{sec:needDA-virtualDPM}

From the very beginning of our work, it was clear that there is a {\em domain gap} between virtual and real worlds. However, it was also clear that this was the case when using images coming from different (real) camera sensors and environments. In other words, the domain gap is not a virtual-to-real issue, but rather a more general sensor-to-sensor or environment-to-environment problem \cite{VazquezNIPSDATA11Cool, VazquezICMI11Virtual}. Other researchers confirmed this fact too when addressing related but different visual tasks than ours \cite{TorralbaCVPR11Unbiased}. Since then, training visual models in virtual worlds and applying {\em domain adaptation} techniques for their use in real-world scenarios come hand-by-hand for us. In fact, more authors have followed the approach of performing some explicit step of virtual- to real-world domain adaptation, without being an exhaustive list, the reader can address \cite{LaiIJRR10Object, ChenECCV12Recognizing, SunBMVC14Virtual, PanaredaBustoBMVC15Adaptation} as illustrative examples.  

We showed that virtual- to real-world domain adaptation is possible for holistic models based on the HOG+LPB/Linear-SVM paradigm \cite{VazquezPAMI14Virtual} as well as on the Haar+EOH/AdaBoost one \cite{VazquezMIIVAIS13Interactive}. In the former case, proof-of-concept experiments adapting RGB-style synthetic images to far infrared ones (FIR) reported positive results too \cite{SocarrasICCVVisDA13Adapting}. Moreover, for the {\em \myindex{Deformable Part-based Model}} (DPM) \cite{FelzenszwalbPAMI10Object} we also proposed to use virtual worlds and domain adaptation \cite{XuITS14Learning, XuPAMI14Domain}. In most of the cases we focused on {\em supervised domain adaptation}, \ie a relatively few amount of annotated {\em target-domain} data (\ie from the real world in our case) was used to adapt the model learned with {\em source-domain} data (from the virtual world). For the holistic models we focused on mixing the source and target data collected via {\em active learning} for model adaptation, we termed the corresponding feature space as {\em cool world}; while for DPM we focused on using just the source-domain model together with the target-domain data, \ie without revisiting the source-domain data. In terms of modern deep CNNs, the former case would be similar to mixing source and target data in the mini-batches while the latter case is more in the spirit of the so-called fine-tuning. 

In the rest of this chapter we are going to focus on DPM because it was the state-of-the-art for object detection before the breakthrough of deep CNNs. A priori it is a good proxy for deep CNNs regarding the specific experiments we want to address, after all deep CNNs eventually can require domain adaptation too \cite{GaninICML15Unsupervised, TzengICCV15Simultaneous, TommasiGCPR15Deeper, ChuECCVTASKCV16Best}. Obviously, being based on HOG-style features there is a point where much more data would not really translate to better accuracy \cite{ZhuIJCV16Need}, so we will keep training data in the order of a few thousands here. On the other hand, note that DPM can be reformulated as a deep CNN \cite{GirshickCVPR15Deformable} for end-to-end learning. Moreover, the domain adaptation techniques we proposed for DPM \cite{XuPAMI14Domain}, can be used as core technology for hierarchical domain adaptation\footnote{With this technique we won the 1st pedestrian detection challenge of the KITTI benchmark suite, a part of the {\em Recognition Meets Reconstruction Challenge} held in ICCV'13.} \cite{XuIJCV16Hierarchical} as well as for weakly supervised incremental domain adaptation \cite{XuICRA16Hierarchical}. 

In particular, we are going to rely on our domain adaptation method for DPM termed as {\em \myindex{Structure-aware Adaptive Structural SVM}} (SA-SSVM), which gave us the best performance in \cite{XuPAMI14Domain}. In this chapter we compliment the experiments run in \cite{XuPAMI14Domain} mainly by addressing questions such as {\em the role of photo-realism in the virtual world}, as well as {\em how does the domain gap behave in virtual-vs-real data with respect to dominant object appearance per domain}. Moreover, for the sake of analyzing new use cases, instead of focusing on pedestrian detection using virtual data from Half-Life 2 as in \cite{XuPAMI14Domain}, here we focus on vehicle detection using different virtual-world datasets, namely Virtual KITTI \cite{GaidonCVPR16Virtual}, SYNTHIA \cite{RosCVPR16Synthia}, and GTA \cite{RichterECCV16Playing}.

\section{Domain Adaptation for DPM in a Nutshell}
\label{sec:DADPM-virtualDPM}

\begin{figure}
\centering
\includegraphics[width=0.5\textwidth]{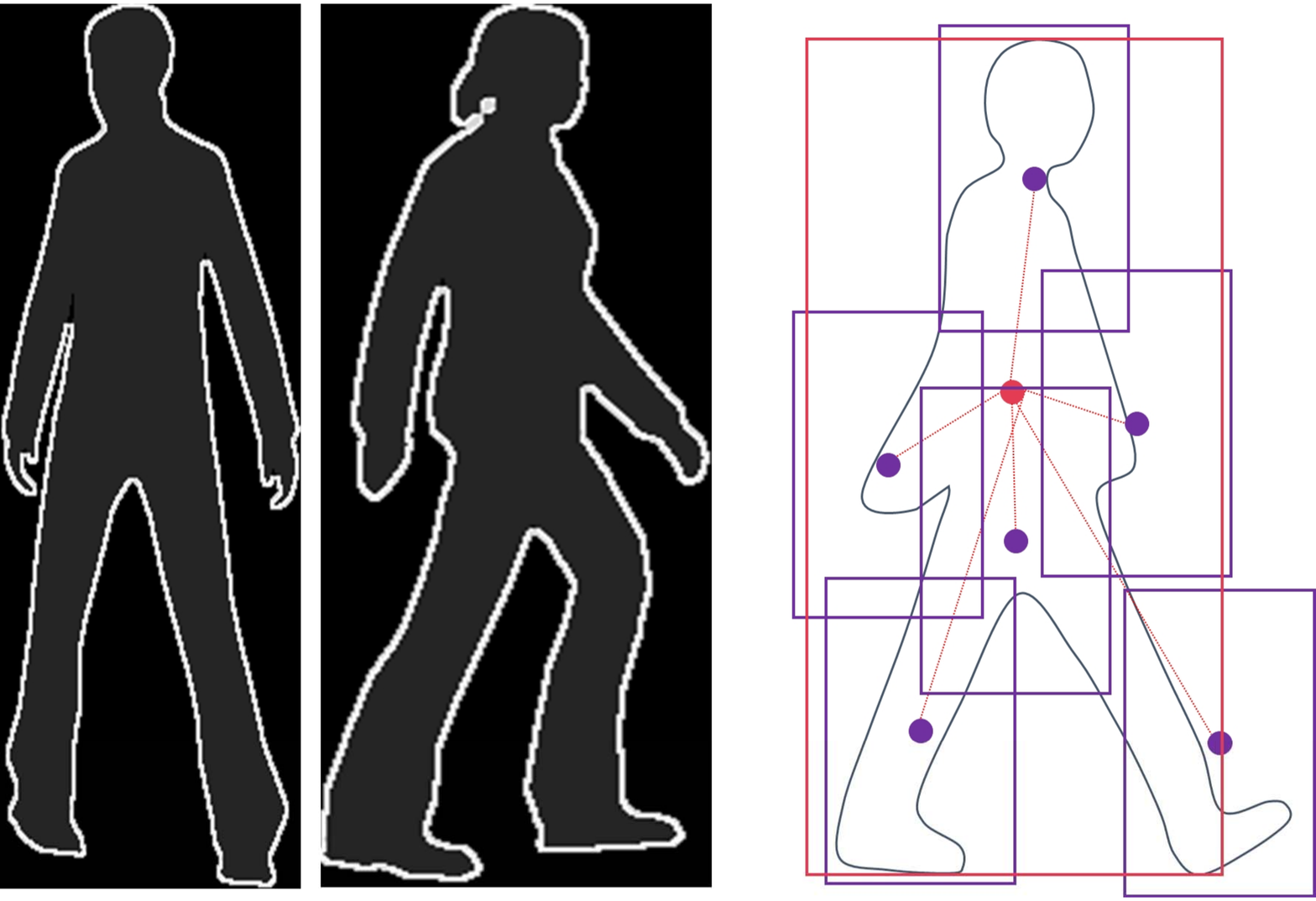}
\caption{DPM for modeling pedestrians. There are two components (left, black background), and each component is encoded as a root and six parts (right, one component).}
\label{fig:Model}
\end{figure}

DPM encodes the {\em appearance} of objects' constituent {\em parts} together with a {\em holistic} object representation termed as {\em root}. In contrast to other part models, DPM allows the parts to be located at different positions with respect to the root. The plausible relative locations, known as {\em deformations}, are also encoded. Both appearance and deformations are learned. The appearance of the parts is learned at double the resolution than the root. The triplet root-parts-deformations is known as {\em component}. In order to avoid too blurred models DPM allows to learn a mixture of components. Different components use to correspond to very different object views or poses, specially when this implies very different aspect ratios of the corresponding root BB. See \fig{Model} for a pictorial intuition. 

\begin{figure*}
\centering
\includegraphics[width=\textwidth]{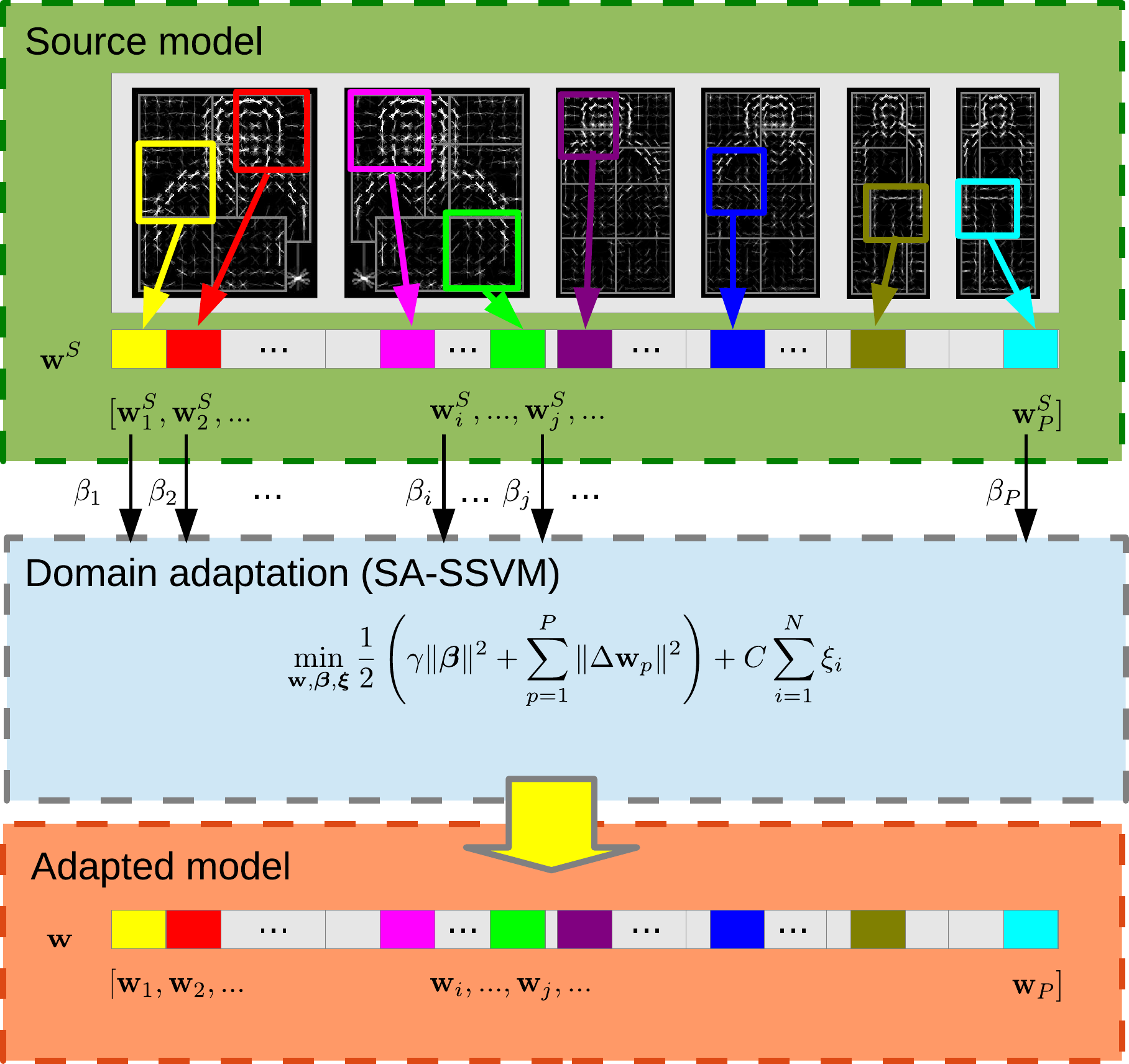}
\caption{Domain Adaptation of DPM based on SA-SSVM (see main text for details).}
\label{fig:SA-SSVM}
\end{figure*}

In practice, a DPM is encoded as a vector $\textbf{w}$ which has to be learned. In the domain adaptation context, we term as $\textbf{w}^S$ the model learned with source-domain data (\eg with virtual-world data). Our SA-SSVM domain adaptation method takes $\textbf{w}^S$ and relatively few annotated target-domain data (\eg real-world data) to learn a new $\textbf{w}$ model which is expected to perform better in the target domain. The reader is referred to \cite{XuPAMI14Domain} for the mathematical technical details of how SA-SSVM works. However, let us explain the idea with the support of the example in \fig{SA-SSVM}; where $\textbf{w}^{S}$ consists of components: half body and full body, as well as persons seen from different viewpoints. Each component consists of root and parts (head, torso, etc). To adapt this DPM to a target domain, we decompose it as $\textbf{w}^{S} = [{\textbf{w}_{1}^{S}}',\dots,{\textbf{w}_{P}^{S}}']'$, where $P$ is the number of structures and $u'$ stands for transpose of $u$. Note that each component, ${\textbf{w}_{p}^{S}}$, may contain both appearance and deformation parameters (for roots only appearance). The decomposed model parameters are adapted to the target domain by different weights, denoted by $\beta_{p}, p \in \{1,P\}$; \ie the SA-SSVM procedure allows domain adaptation for each of such structures separately by defining $\Delta \textbf{w}_{p} = \textbf{w}_{p}-\beta_{p}\textbf{w}^{S}_{p}$, $p\in \{1,P\}$. In order to learn these adaptation weights, we further introduce a regularization term ${\|{\boldsymbol\beta}\|^{2}}$ in the objective function, where $\boldsymbol\beta = [\beta_{1},\dots,\beta_{P}]'$, and we use a scalar parameter $\gamma$ to control its relative penalty. Finally, $C$ and the $\xi_i$ are just the standard terms of a SVM objective function and $N$ the number of target-domain samples used for the adaptation. After optimizing for the objective function (see mid box in \fig{SA-SSVM}), $\textbf{w} = [{\textbf{w}_{1}}',\dots,{\textbf{w}_{P}}']'$ is the domain adapted DPM.

\section{Experimental Results}
\label{sec:experiments-virtualDPM}

\subsection{Datasets}
\label{ssec:datasets-virtualDPM}

\begin{figure*}
\includegraphics[width=\textwidth]{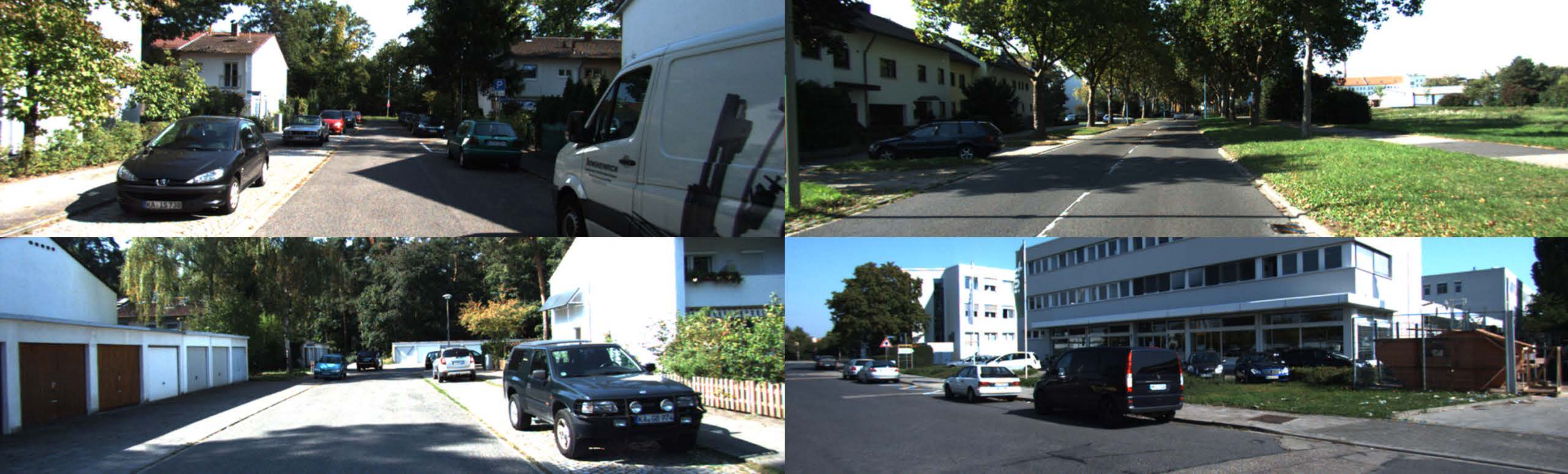}
\caption{Images sampled from KITTI-Det.}
\label{fig:KITTI-Detect}
\end{figure*}

\begin{figure*}
\includegraphics[width=\textwidth]{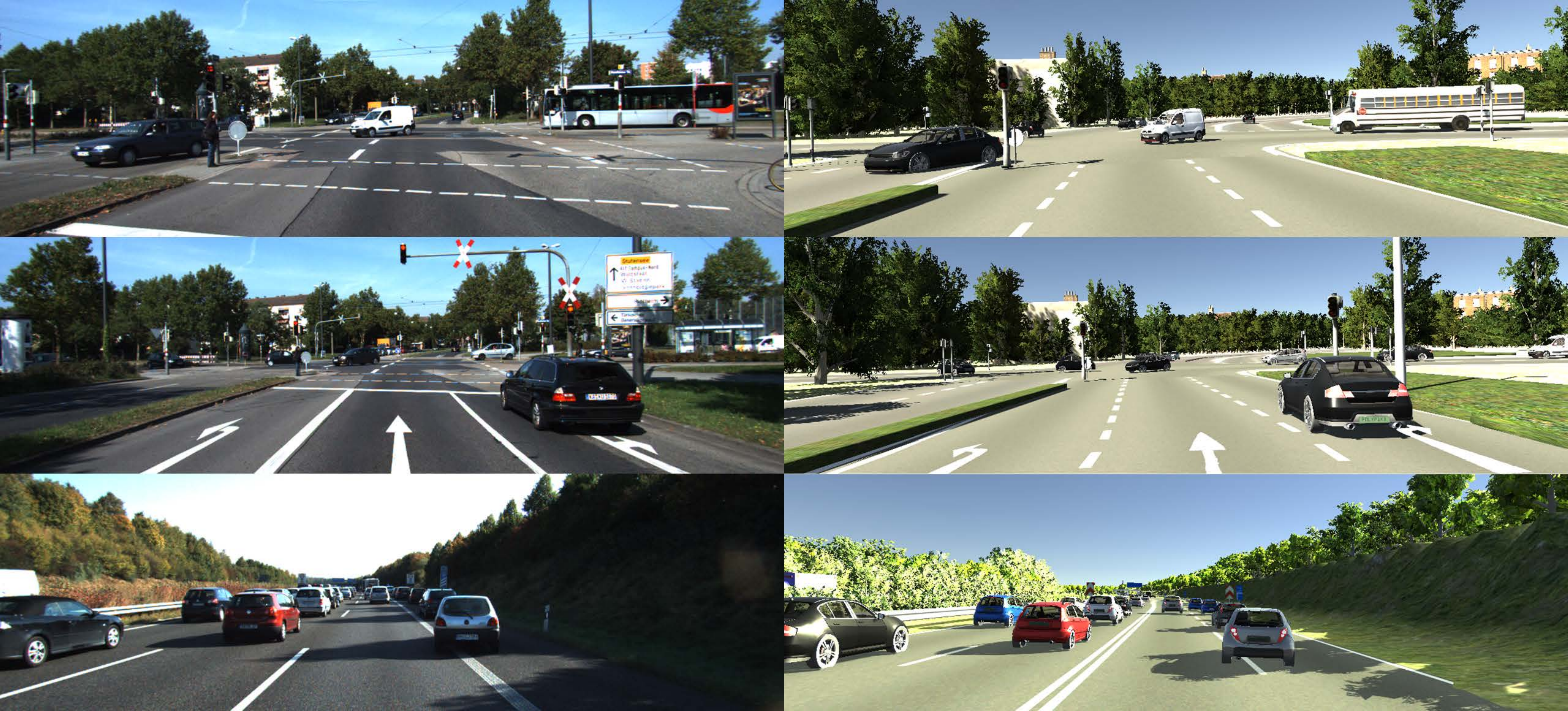}
\caption{Images sampled from KITTI-Track (left) and Virtual KITTI (right). Note how Virtual KITTI is a synthesized but realistic clone of KITTI-Track.}
\label{fig:KITTI-Track}
\end{figure*}

\begin{figure*}[t]
\includegraphics[width=\textwidth]{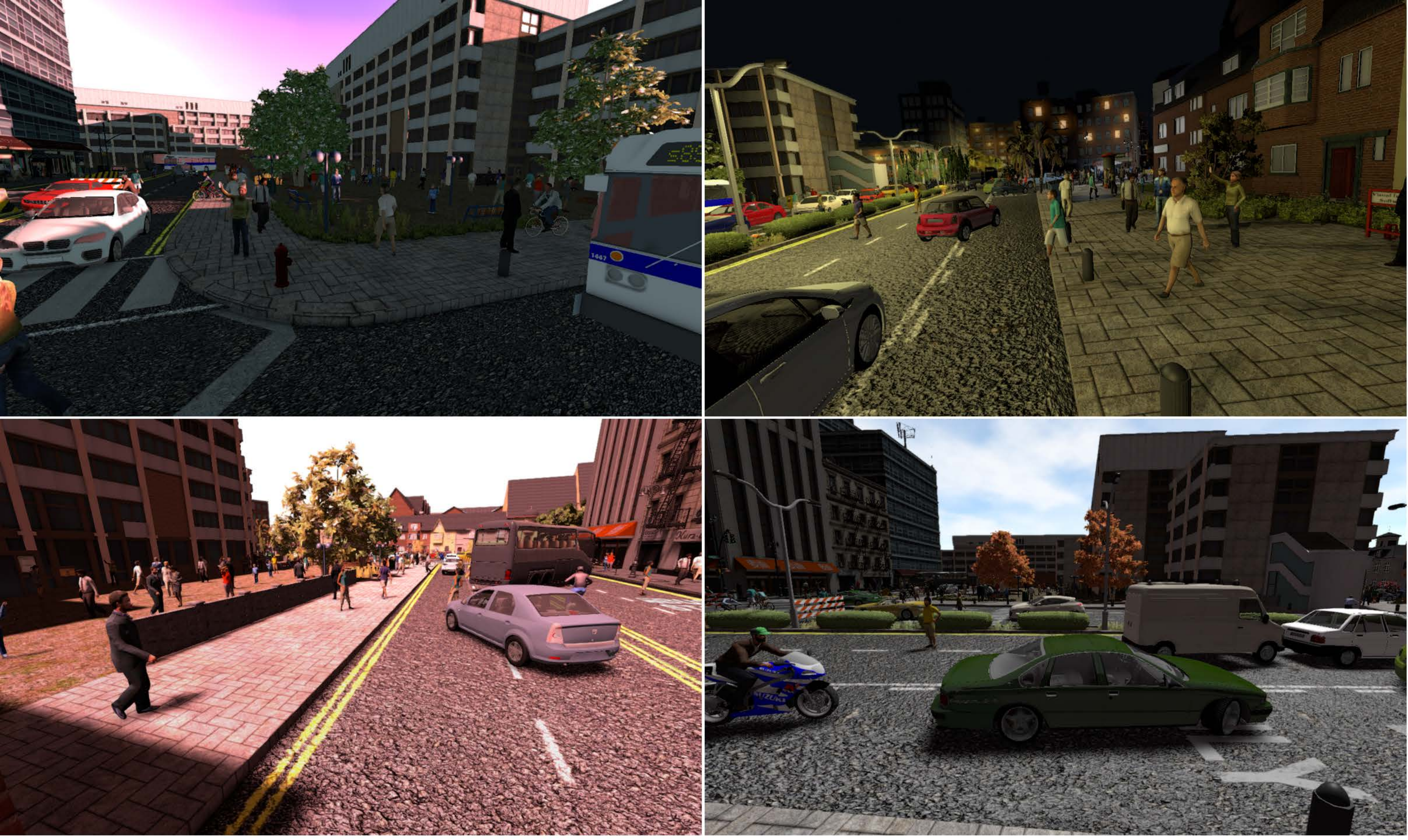}
\caption{Images sampled from SYNTHIA dataset. Note that they are not always corresponding to a forward facing virtual camera on-board a car.}
\label{fig:SYNTHIA}
\end{figure*}

\begin{figure*}
\includegraphics[width=\textwidth]{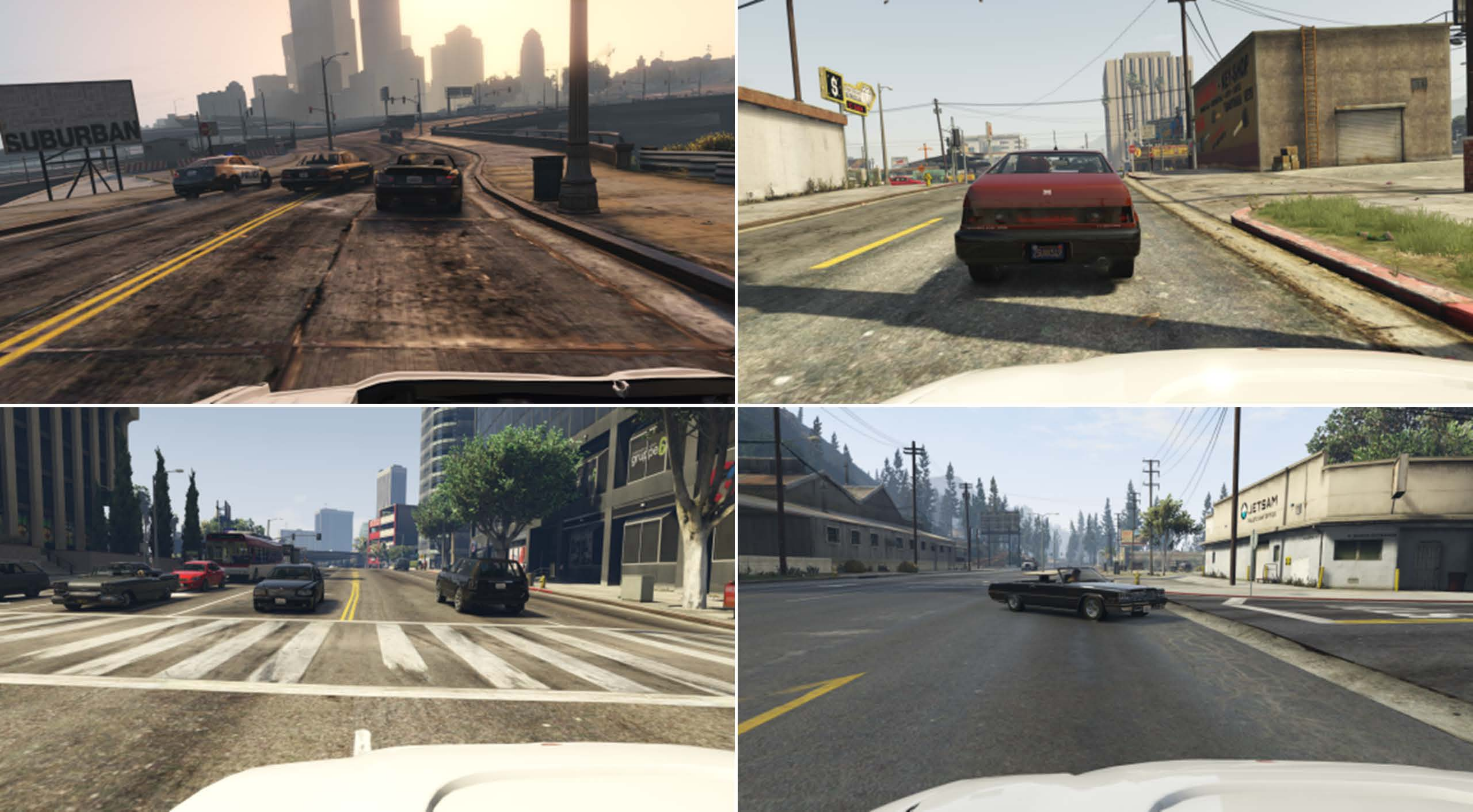}
\caption{Images sampled from the GTA videogame.}
\label{fig:GTA}
\end{figure*}

As we mentioned before, we are going to focus on vehicle detection for ADAS/AD applications. We use the training data of the KITTI car detection challenge \cite{GeigerCVPR16Ready}; which is split into two sets, one for actually training and the other for testing. Such a testing set will be the only one used here for that purpose, and we will follow the so-called {\em moderate} setting when considering which vehicles are mandatory to detect. For training, we will consider four more datasets in addition to the mentioned split of the KITTI car detection challenge, thus, five in total. Namely, the KITTI car tracking dataset \cite{GeigerCVPR16Ready}, its synthesized clone \myindex{Virtual KITTI} \cite{GaidonCVPR16Virtual}, SYNTHIA \cite{RosCVPR16Synthia}, and \myindex{GTA} \cite{RichterECCV16Playing}. Of course, SYNTHIA is a dataset with different types of ground truth, so we selected a set of cars and images for our experiments. In the case of GTA, we semi-automatically annotated with BBs a set of cars. Table \ref{tab:numberSamples} shows the number of samples in each dataset. Figures \ref{fig:KITTI-Detect}, \ref{fig:KITTI-Track}, \ref{fig:SYNTHIA}, and \ref{fig:GTA}, show images sampled from KITTI-Det, KITTI Track with Virtual KITTI, SYNTHIA and GTA, respectively. Virtual KITTI and SYNTHIA are based on the same development framework, \ie Unity3D\footnote{See \tt{unity3d.com}}. We can see that the images from GTA are more photo-realistic than the ones used in SYNTHIA and Virtual KITTI. SYNTHIA images are not always corresponding to a forward facing on-board virtual camera as is the case of Virtual KITTI and GTA. For more details about the datasets the reader can refer to the corresponding papers.


\begin{table*}
\caption{Used samples for each dataset. {\em Images} stands for the number of images and, from them, {\em Vehicles} stands for the number of annotated vehicles using a bounding box. Negative samples are selected from background areas of the same images. {\em KITTI-Det Test} and {\em KITTI-Det Train} refer to two splits of the training set of the KITTI car detection training set. {\em KITTI-Det Test} is the testing set used in all the experiments of this chapter, while the rest of datasets are used only for training. For {\em KITTI Track} and {\em Virtual KITTI}, we use sequences 1, 2, 6, 18, and 20 as the training datasets. {\em SYNTHIA-sub} refers to a subset randomly sampled from {\em SYNTHIA}.}
\begin{scriptsize}
\begin{center}
\begin{tabular}{|l||c||c|c|c|c|c|c|}
\hline
                 & KITTI-Det Test & KITTI-Det Train & KITTI-Track & Virtual KITTI & SYNTHIA & SYNTHIA-Sub & GTA\\ \hline
 Images          &3163  &3164  &2020  &1880 &1313 &675   &580\\ \hline
 Vehicles        &12894 &12275 &12950 &6867 &2052 &1023  &1054\\ \hline
\end{tabular}
\end{center}
\end{scriptsize}
\label{tab:numberSamples}
\end{table*}

\subsection{Protocol}
\label{ssec:protocol-virtualDPM}

In order to show the accuracy of the vehicle detectors we plot curves of {\em false positive per image} (FPPI) \vs {\em miss rate} (MR) according to the Caltech protocol \cite{DollarPAMI12Pedestrian}; with an overlap of the $50\%$ between detection and ground truth BBs. For training and testing we only consider {\em moderate} cases, which according to the definition given in the KITTI car detection challenge, are those vehicles non-occluded or just partially occluded (maximum truncation: $30\%$), and with a BB height $\geq25$ pixels. The vehicles mentioned in \tab{numberSamples} refer to moderate cases.

Regarding DPM we use three components, each with eight parts. Part locations are initialized as $6\times6$ HOG-style cells ($48\times48$ pixels) covering the root (at its double resolution version). Note that, in contrast to \cite{XuITS14Learning, XuPAMI14Domain}, here we have not explored the use of the pixel-wise vehicle masks (available for virtual-world data) to provide a better initialization of part locations during DPM learning. Thus, real- and virtual-world training data are used equally for learning source-domain DPMs.

Regarding the application of SA-SSVM we have followed the settings reported in \cite{XuPAMI14Domain} as producing the best results. Namely, the adapted structures correspond to the root and parts, \ie not to components; and we set $\gamma=0.08$ and $C=0.001$ (see \fig{SA-SSVM}). Since domain adaptation experiments (\ie SA-SSVM based ones) require random sampling of the target domain training data, they are run three times and the mean FPPI-MR curve and standard-deviation based intervals are plotted (\ie as in \cite{XuPAMI14Domain} but with three repetitions instead of five).

\subsection{Experiments}
\label{ssec:experiments-virtualDPM}

According to the datasets listed in \tab{numberSamples}, we define the set of source-domain datasets to be $\mathcal{S} = \{$KITTI-Track, Virtual KITTI, SYNTHIA, SYNTHIA-Sub, GTA$\}$. The target-domain dataset is always KITTI-Det Test. KITTI-Det Train and KITTI-Det Test are coming from the same domain since they correspond to two splits we have done from the same original dataset. All the learned detectors are tested in KITTI-Det Test, and the difference among them is the data used for their training.  Accordingly, the reported experiments are as follows:
\begin{itemize}
\item {\em SRC}: Training with a dataset $s\in\mathcal{S}$.
\item {\em TAR-ALL}: Training based on the full KITTI-Det Train.
\item {\em TARX}: Training with a subset of random images from KITTI-Det Train, in particular, only using the $100X\%$ of the images.
\item {\em SA-SSVM}: Training with a dataset $s\in\mathcal{S}$ plus the images used for the {\em TARX} shown in the same plot. 
\end{itemize}

\begin{figure*}
\centering
\subfigure[KITTI-Track]{\includegraphics[width=0.45\textwidth]{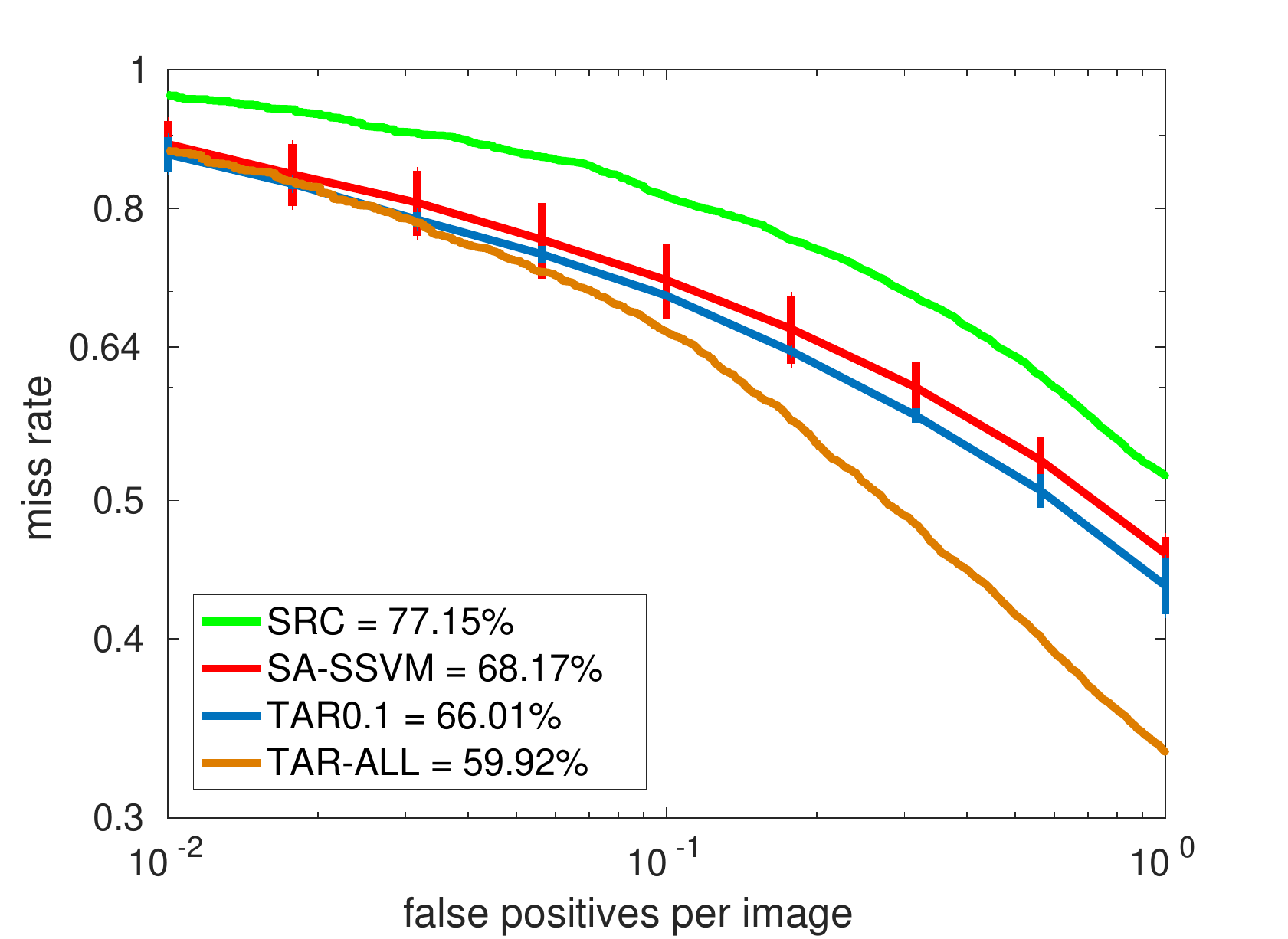}}
\subfigure[Virtual KITTI]{\includegraphics[width=0.45\textwidth]{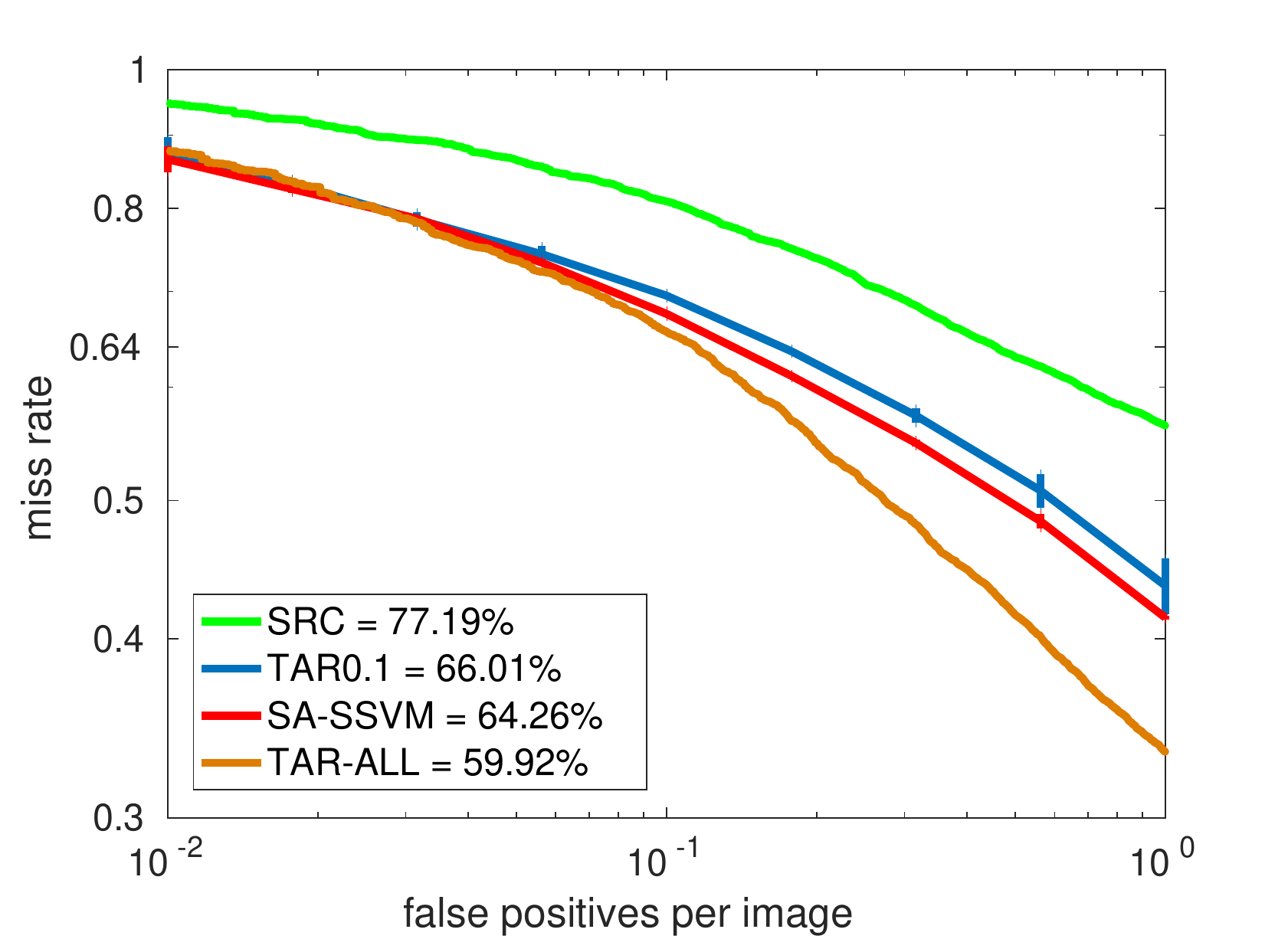}}
\subfigure[SYNTHIA]{\includegraphics[width=0.45\textwidth]{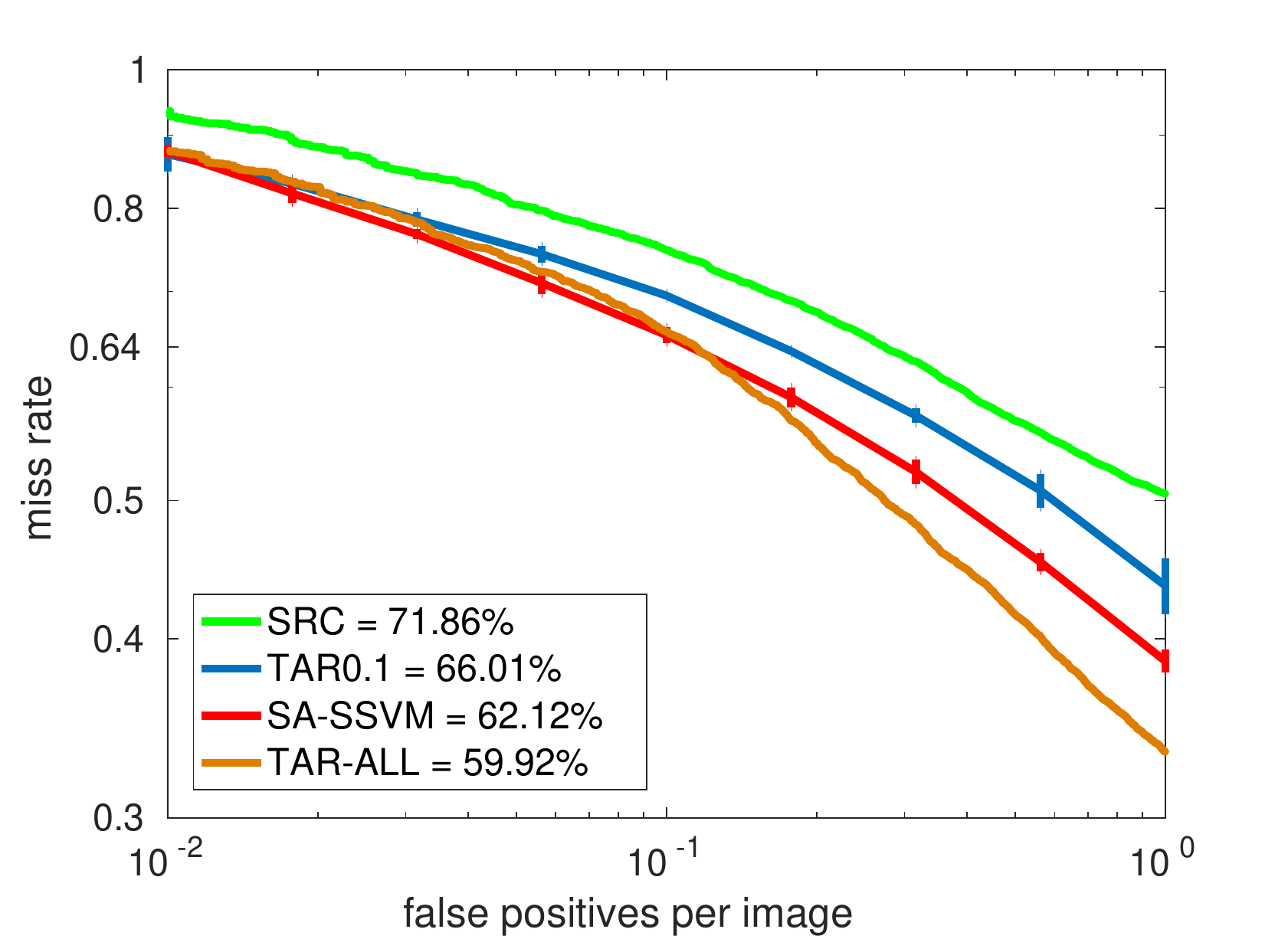}}
\subfigure[SYNTHIA-Sub]{\includegraphics[width=0.45\textwidth]{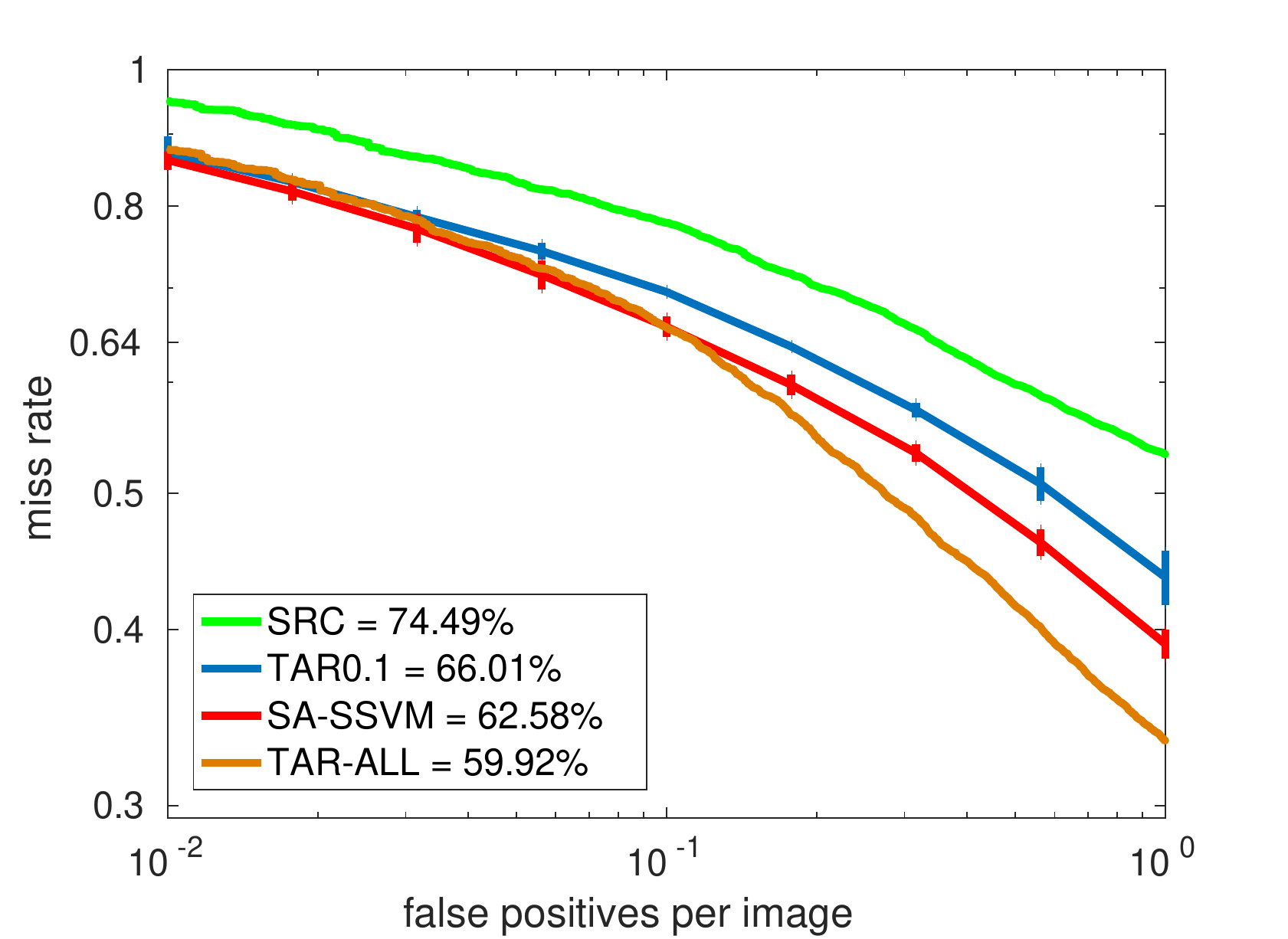}}
\subfigure[GTA]{\includegraphics[width=0.45\textwidth]{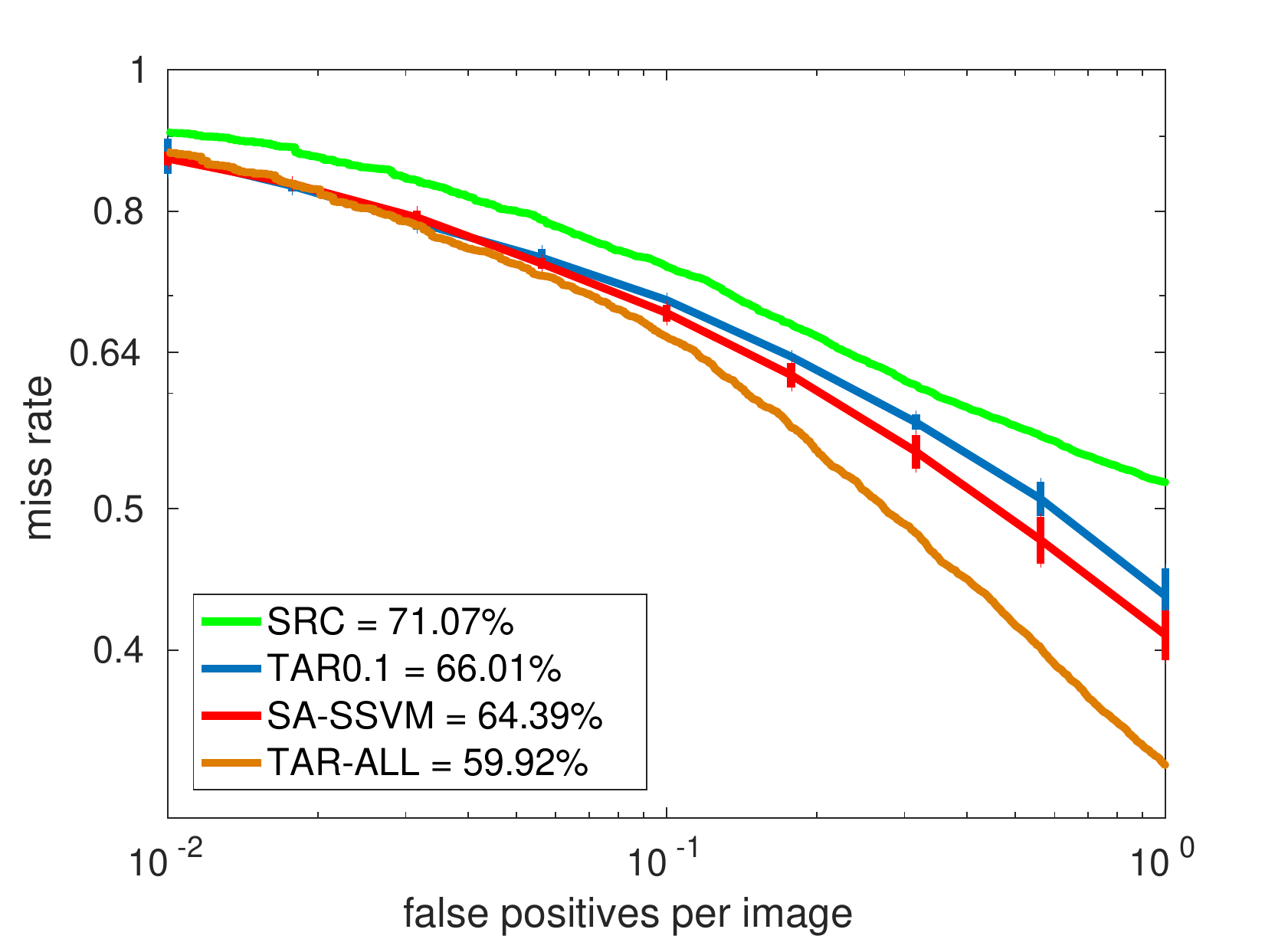}}
\caption{Results assuming $X=0.1$ (see main text). In the box legend it is indicated the average miss rate for each experiment. Thus, the lower the better.}
\label{fig:plotsX10}
\end{figure*}

\begin{figure*}
\centering
\subfigure[KITTI-Track]{\includegraphics[width=0.45\textwidth]{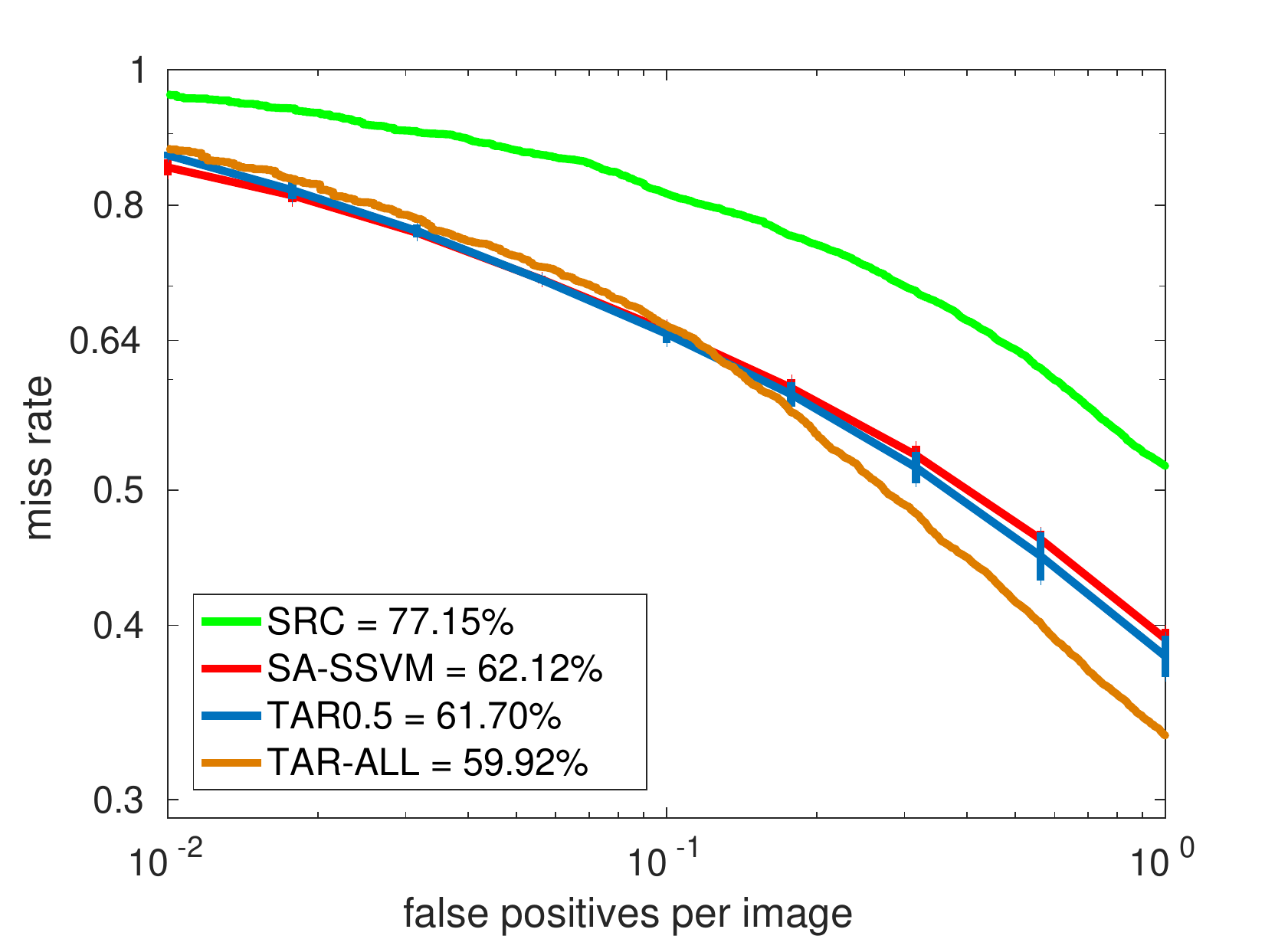}}
\subfigure[Virtual KITTI]{\includegraphics[width=0.45\textwidth]{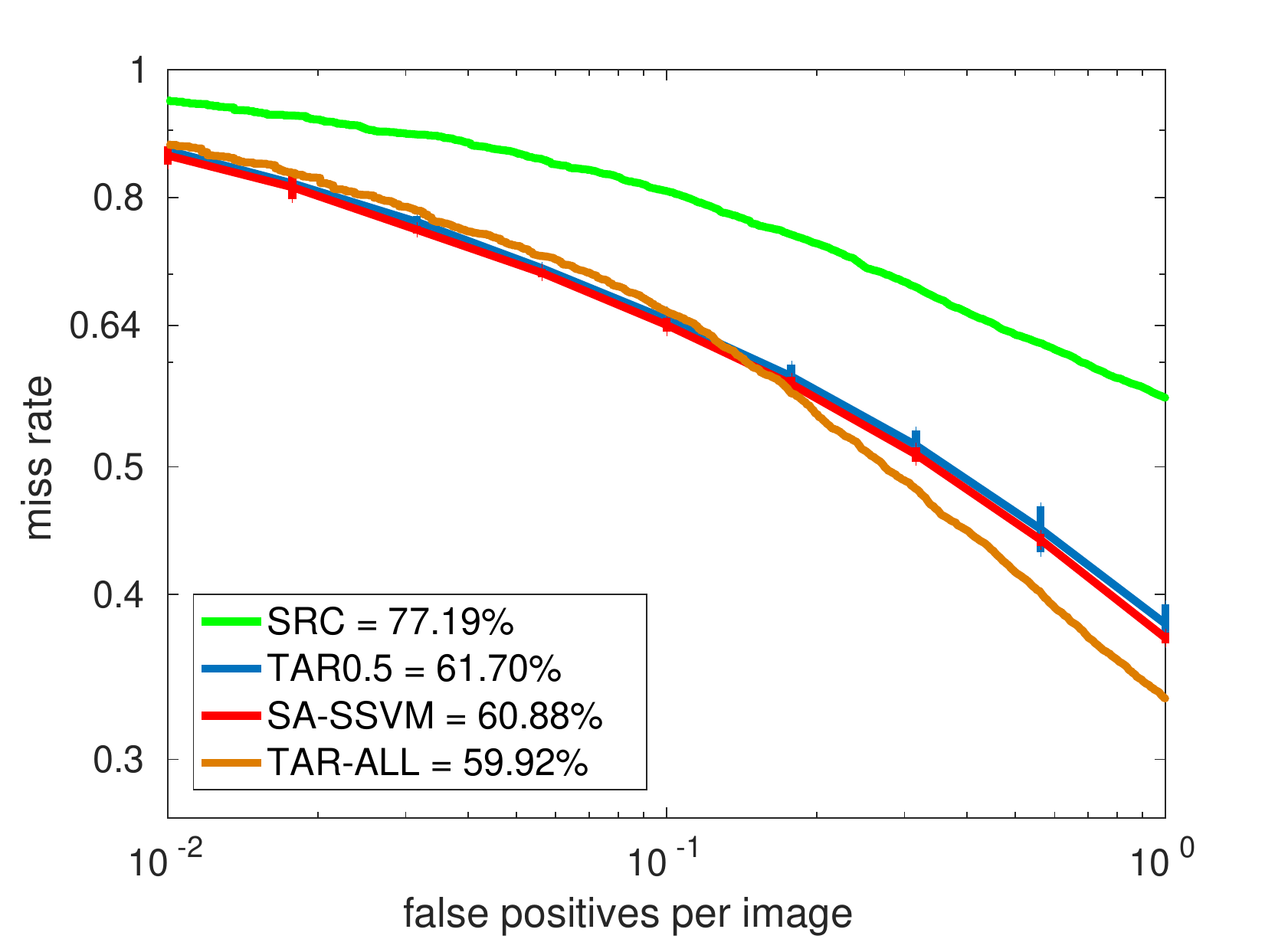}}
\subfigure[SYNTHIA]{\includegraphics[width=0.45\textwidth]{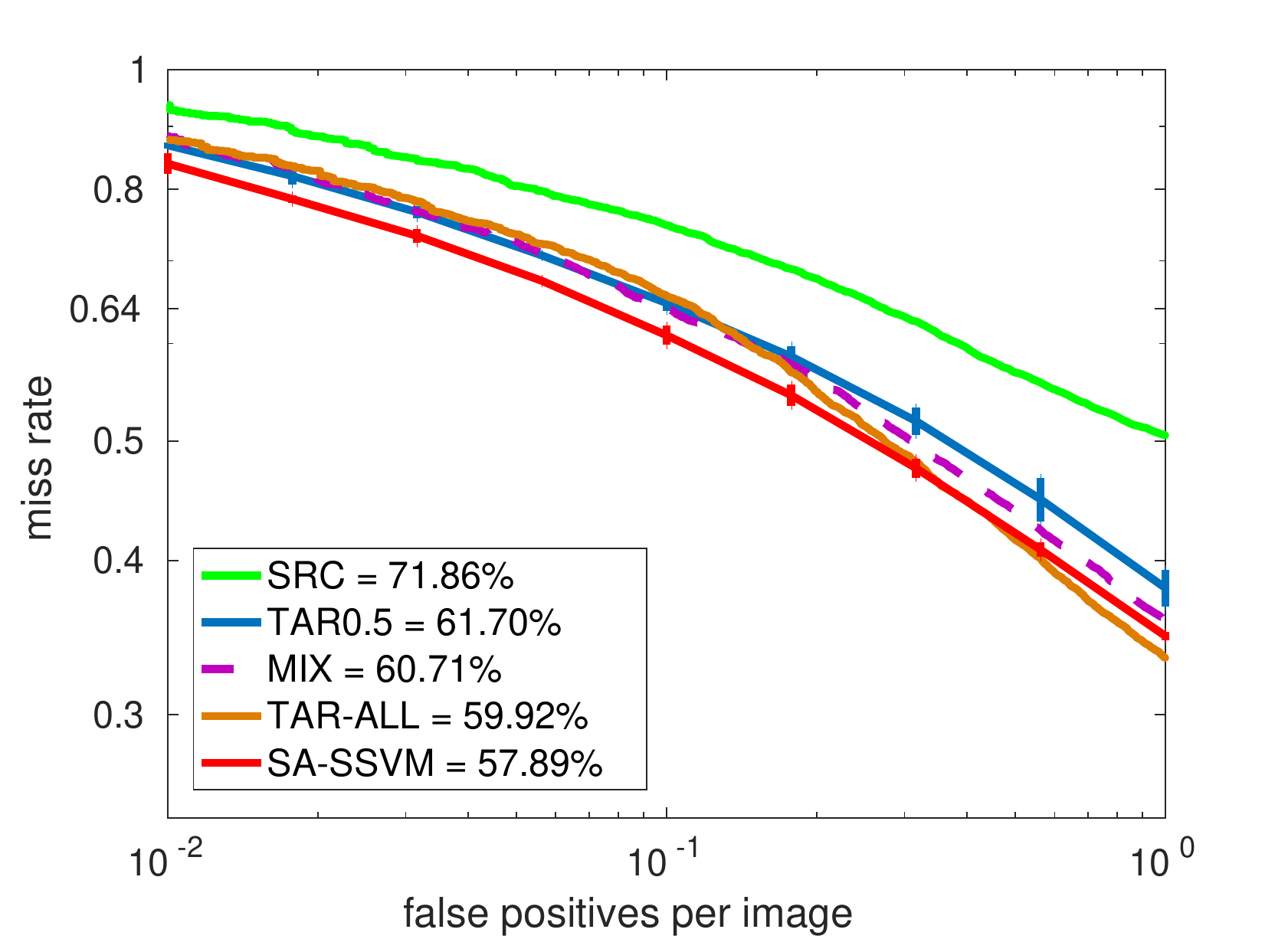}}
\subfigure[SYNTHIA-Sub]{\includegraphics[width=0.45\textwidth]{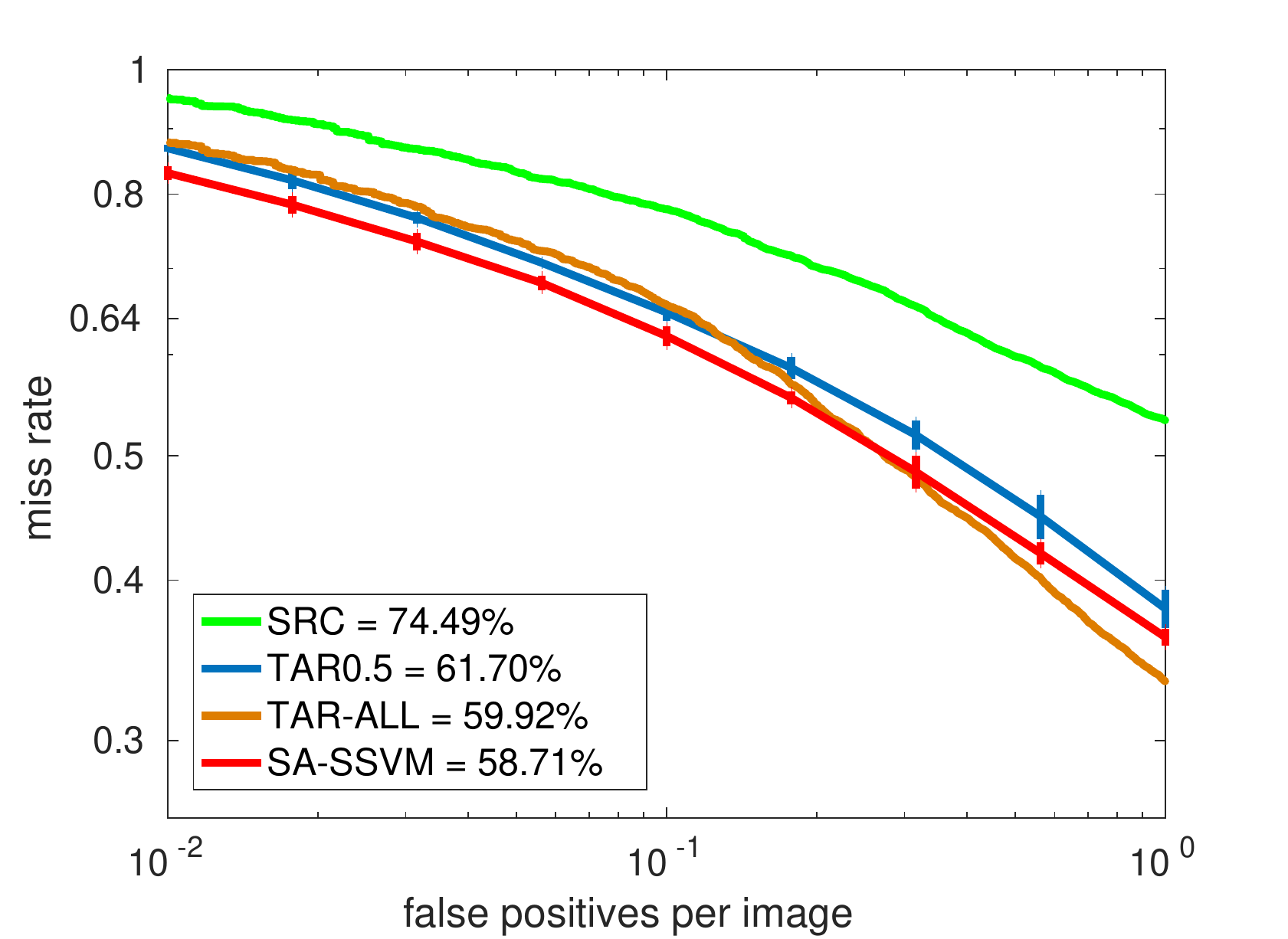}}
\subfigure[GTA]{\includegraphics[width=0.45\textwidth]{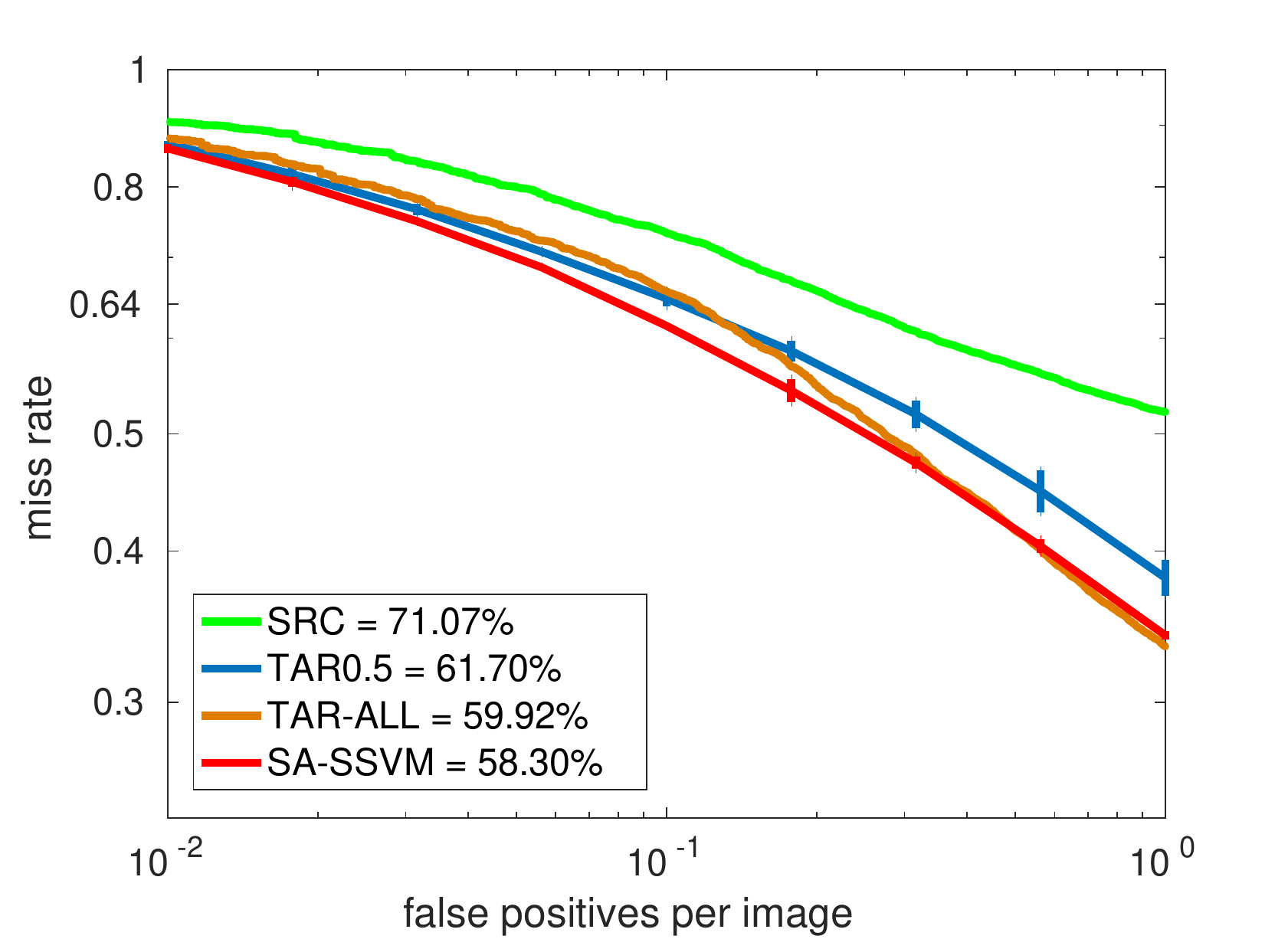}}
\caption{Results assuming $X=0.5$ (see main text). In the box legend it is indicated the average miss rate for each experiment. Thus, the lower the better.}
\label{fig:plotsX50}
\end{figure*}

\begin{figure*}
\centering
\subfigure[KITTI-Track]{\includegraphics[width=0.45\textwidth]{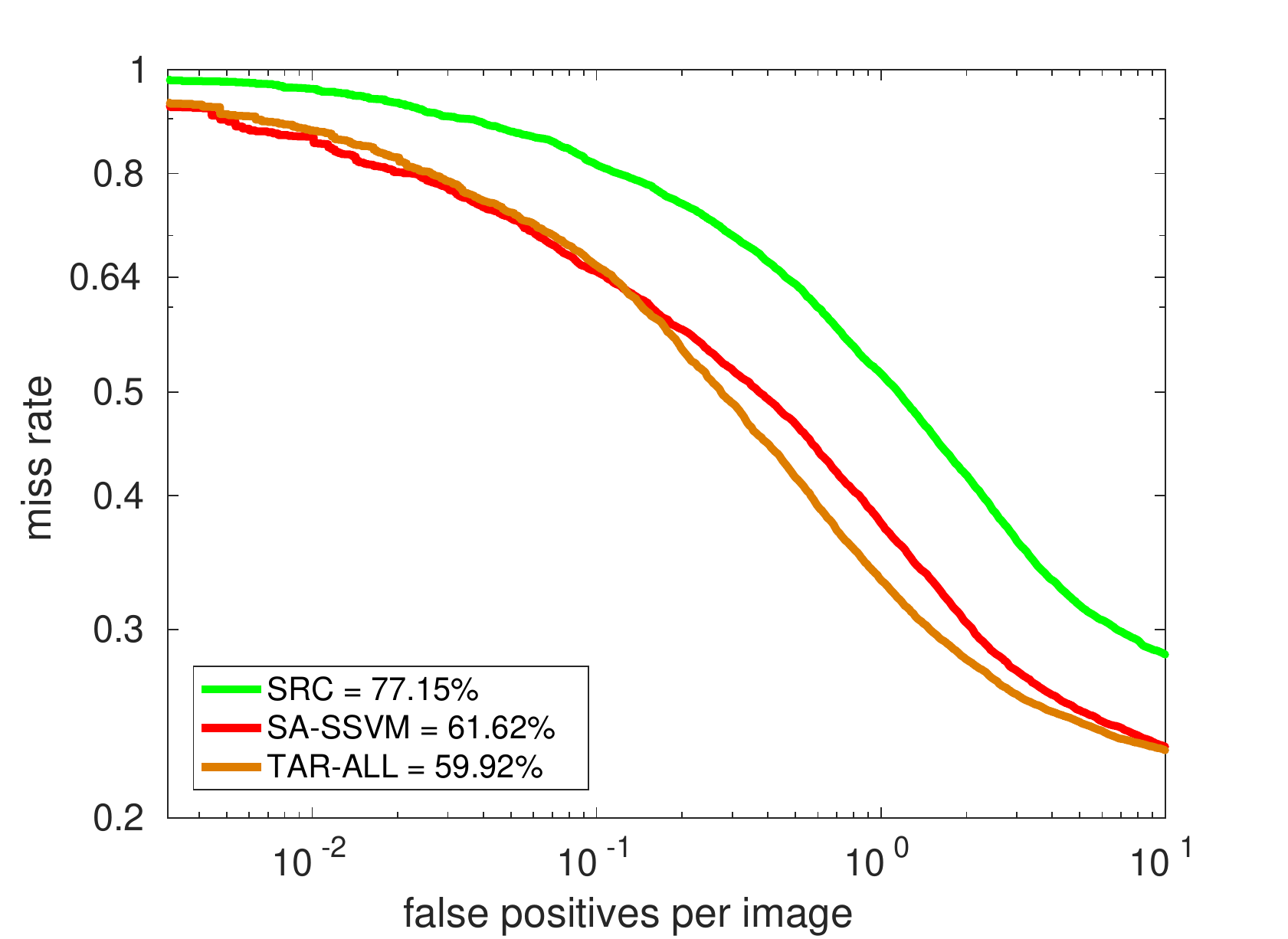}}
\subfigure[Virtual KITTI]{\includegraphics[width=0.45\textwidth]{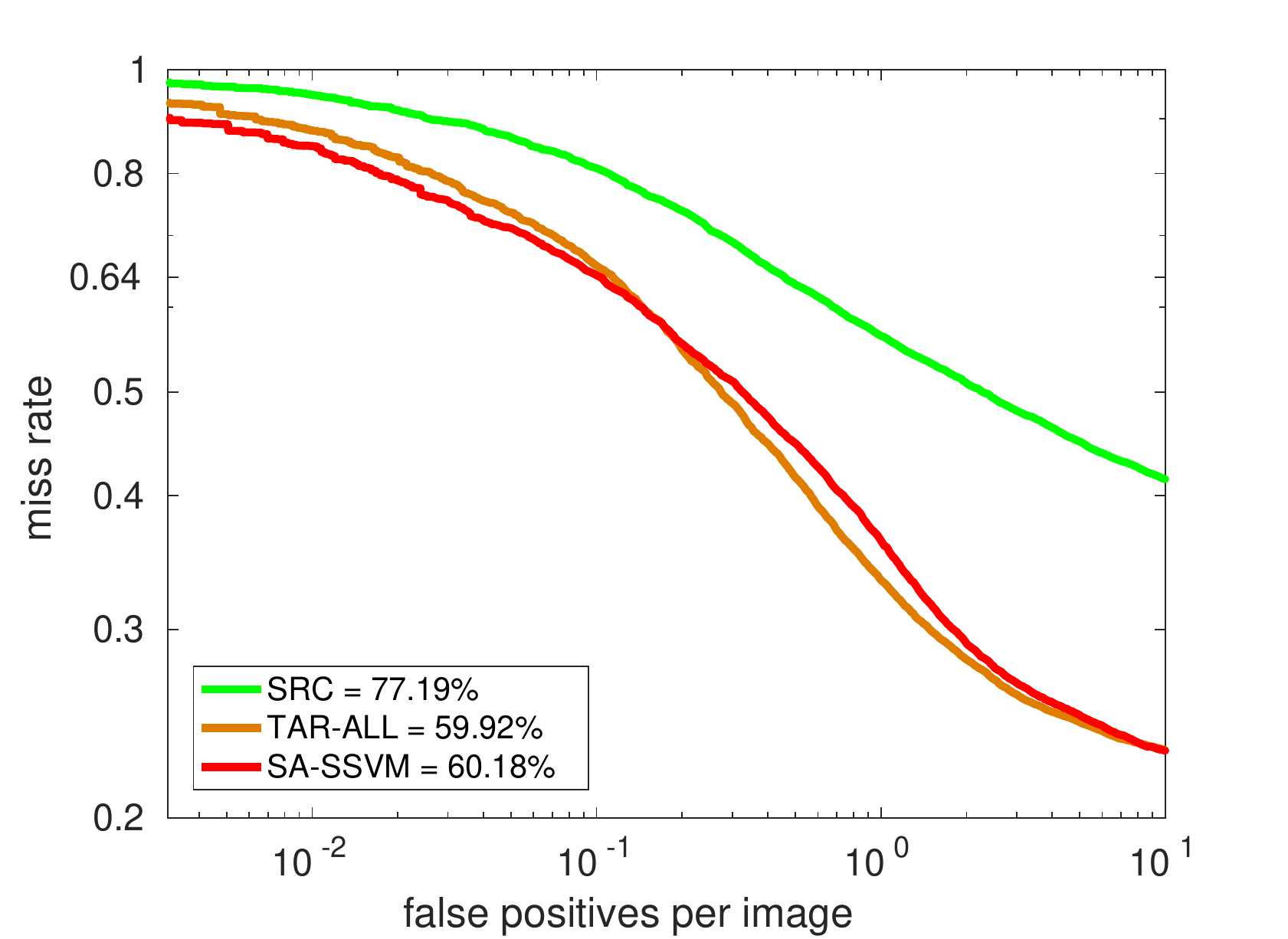}}
\subfigure[SYNTHIA]{\includegraphics[width=0.45\textwidth]{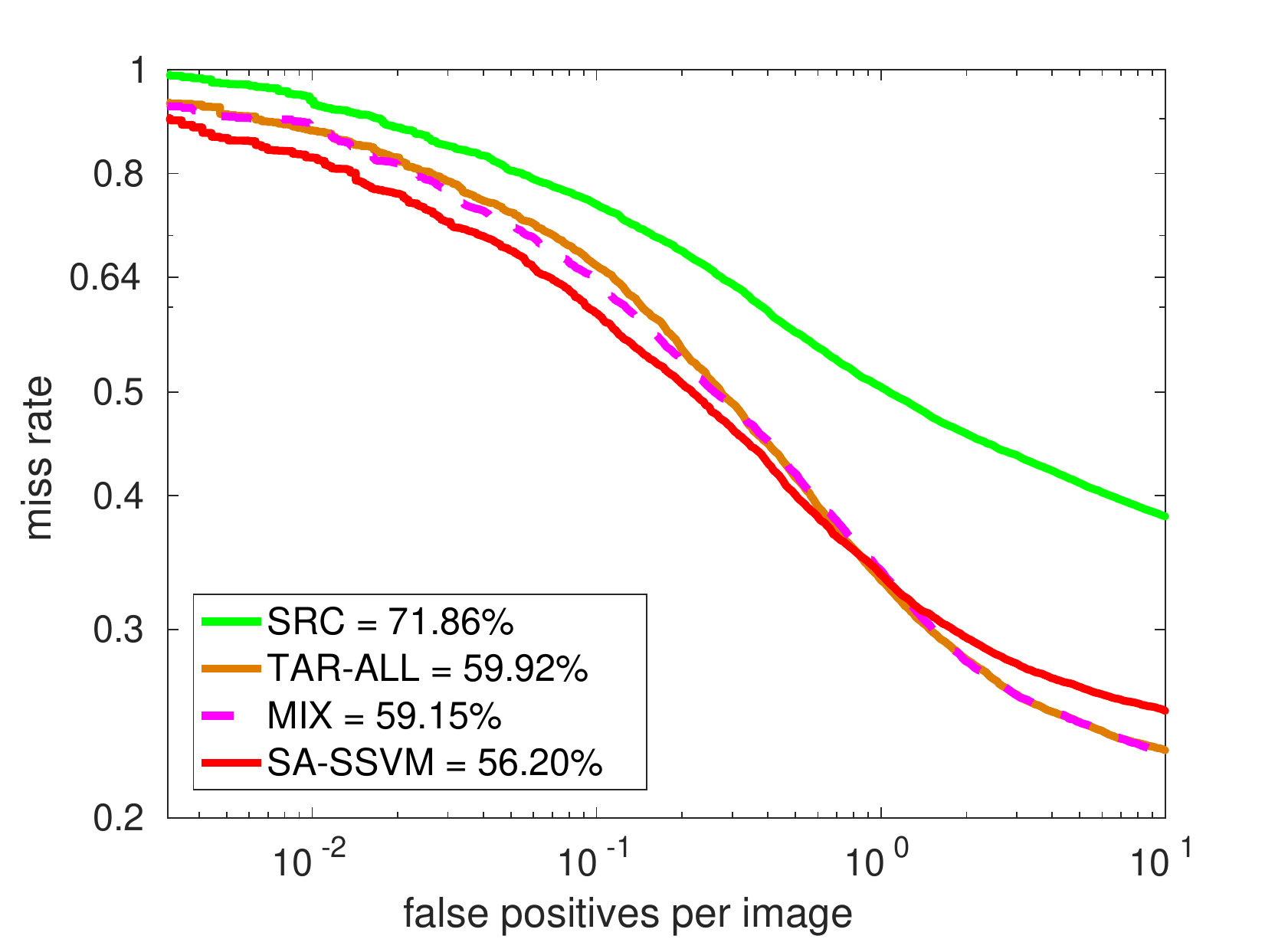}}
\subfigure[SYNTHIA-Sub]{\includegraphics[width=0.45\textwidth]{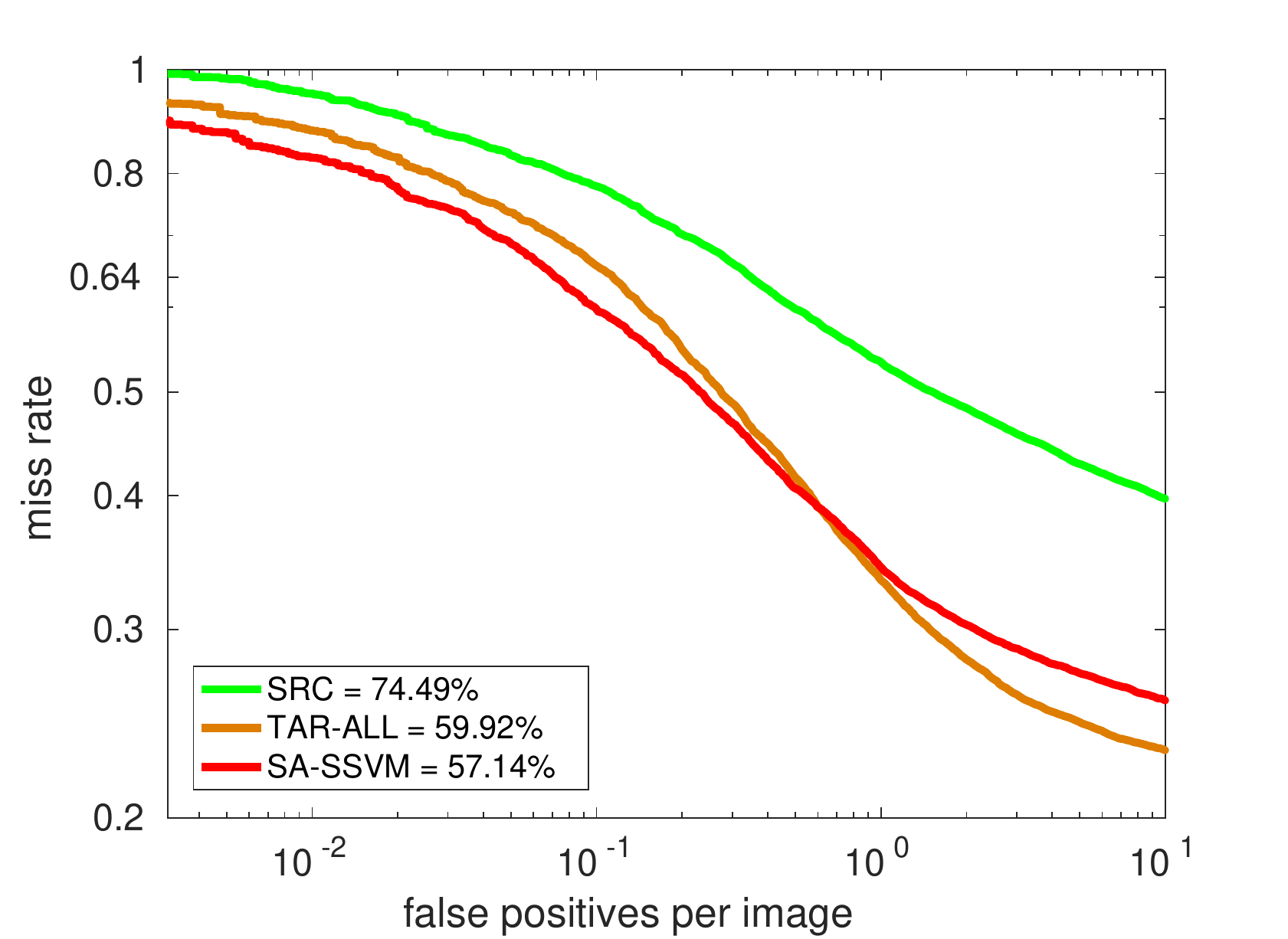}}
\subfigure[GTA]{\includegraphics[width=0.45\textwidth]{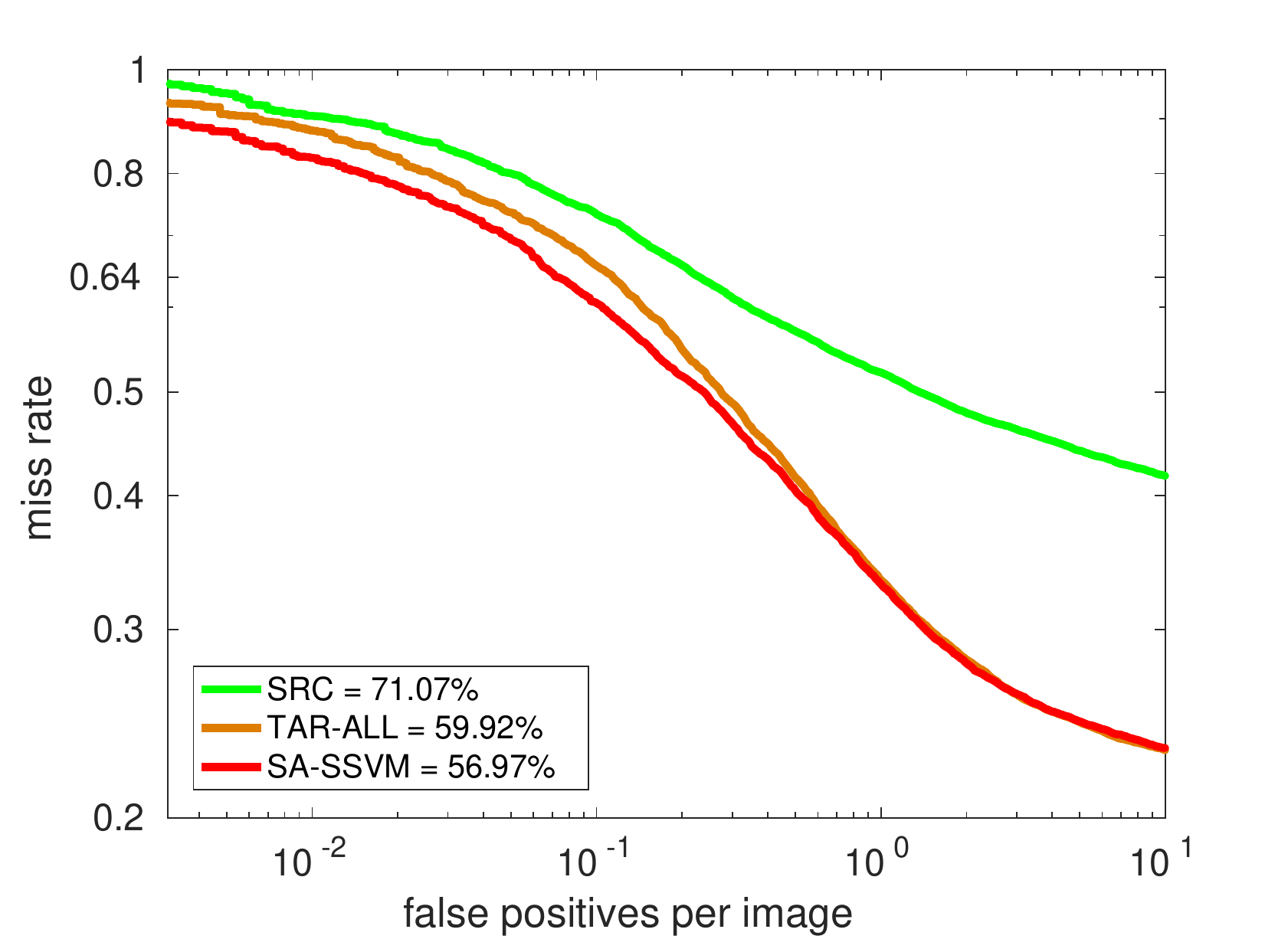}}
\caption{Results assuming $X=1$ (ALL; see main text). In the box legend it is indicated the average miss rate for each experiment. Thus, the lower the better.}
\label{fig:plotsX100}
\end{figure*}

Following this pattern, \fig{plotsX10} shows results for $X=0.1$ (\ie $10\%$), \fig{plotsX50} shows results for $X=0.5$ (\ie $50\%$), and \fig{plotsX100} shows results for $X=1$ (\ie {\em ALL}).

\subsection{Discussion}
\label{ssec:discussion-virtualDPM}

The first observation comparing SRC and TAR-ALL results (note that they are constant across figures since do not depend on $X$) is that there is a large domain gap. Since we would like to annotate as less real-world data as possible, let us start our analysis for $X=0.1$ (see \fig{plotsX10}).

The worst case is when SRC$\in\{$KITTI-Track, Virtual KITTI$\}$ since the average miss rate is $\sim17$ points worse than for TAR-ALL. The best case is when SRC$\in\{$SYNTHIA, GTA$\}$, where the gap is of $\sim12$ points. Note that this is not related to the number of vehicle samples since GTA has $\sim1/6$ of vehicles than Virtual KITTI for training, SYNTHIA $\sim1/3$ than Virtual KITTI, and Virtual KITTI in turn contains $\sim1/2$ of the vehicles in KITTI-Track and in KITTI-Det Test. An analogous behavior is seen when ranking the datasets by number of vehicle-free images. In any case, both $\sim17$ and $\sim12$ points are significant accuracy drops. 

For going deeper in the analysis of what can be the reason for the domain gap, we can compare the results of KITTI-Track \vs Virtual KITTI. We see that they are basically equal. Since Virtual KITTI is a synthesized clone of KITTI-Track, we think that the main reason of the accuracy drop is not the virtual-to-real nature of the training images, but the typical vehicle poses and backgrounds reflected in the training datasets, \ie when comparing Virtual KITTI/KITTI-Track with KITTI-Det Train. In other words, KITTI-Det Train represents better KITTI-Det Test since we built them from the same data set. Note, in addition, that KITTI-Track come from the same camera sensor as KITTI-Det Train and Test, which does not avoid the accuracy gap. Moreover, both SYNTHIA and GTA come from virtual worlds and still produce a detector that performs better than when using KITTI-Track. 

We observe also that leaving out the $\sim90\%$ of the images in KITTI-Det Train ($X=0.1$) causes a drop in accuracy of $\sim6$ points. In other words, in this case once we have $\sim316$ manually annotated images ($\sim10\%$ of KITTI-Det Train), annotating $\sim2,848$ more is required to push the DPM to its limit, which is only $\sim6$ points better\footnote{{\em It is a fallacy to believe that, because good
datasets are big, then big datasets are good} \cite{BergProcIEEE10All}.}. Active learning or ad hoc heuristics can be tried to alleviate such manual annotation effort. However, we observe that pre-training the DPM with automatically collected virtual-word data and using SA-SSVM for adapting the model, makes such $\sim316$ images already very valuable, since in all cases the domain-adapted vehicle detector improves both the result of TAR0.1 and SRC (only virtual-word data). We can see that the best case is for SYNTHIA, which reduces the domain gap to $\sim2$ points from $\sim12$ points, and improves the result of TAR0.1 in $\sim4$ points. An additional observation is that pre-training (SRC) the detectors with virtual-world data also allows to use active learning techniques as we did in \cite{VazquezPAMI14Virtual} and/or ad hoc heuristics as we did in \cite{XuICRA16Hierarchical} for annotating more complementary images (\ie other more informative $\sim316$ ones) or collecting more but without human intervention (\ie self-annotation). We have not done it here for the sake of simplicity, but it is reasonable to think that this would reduce the domain gap even more.

\begin{figure*}
\centering
\includegraphics[width=\textwidth, trim={0, 170, 0, 0}, clip]{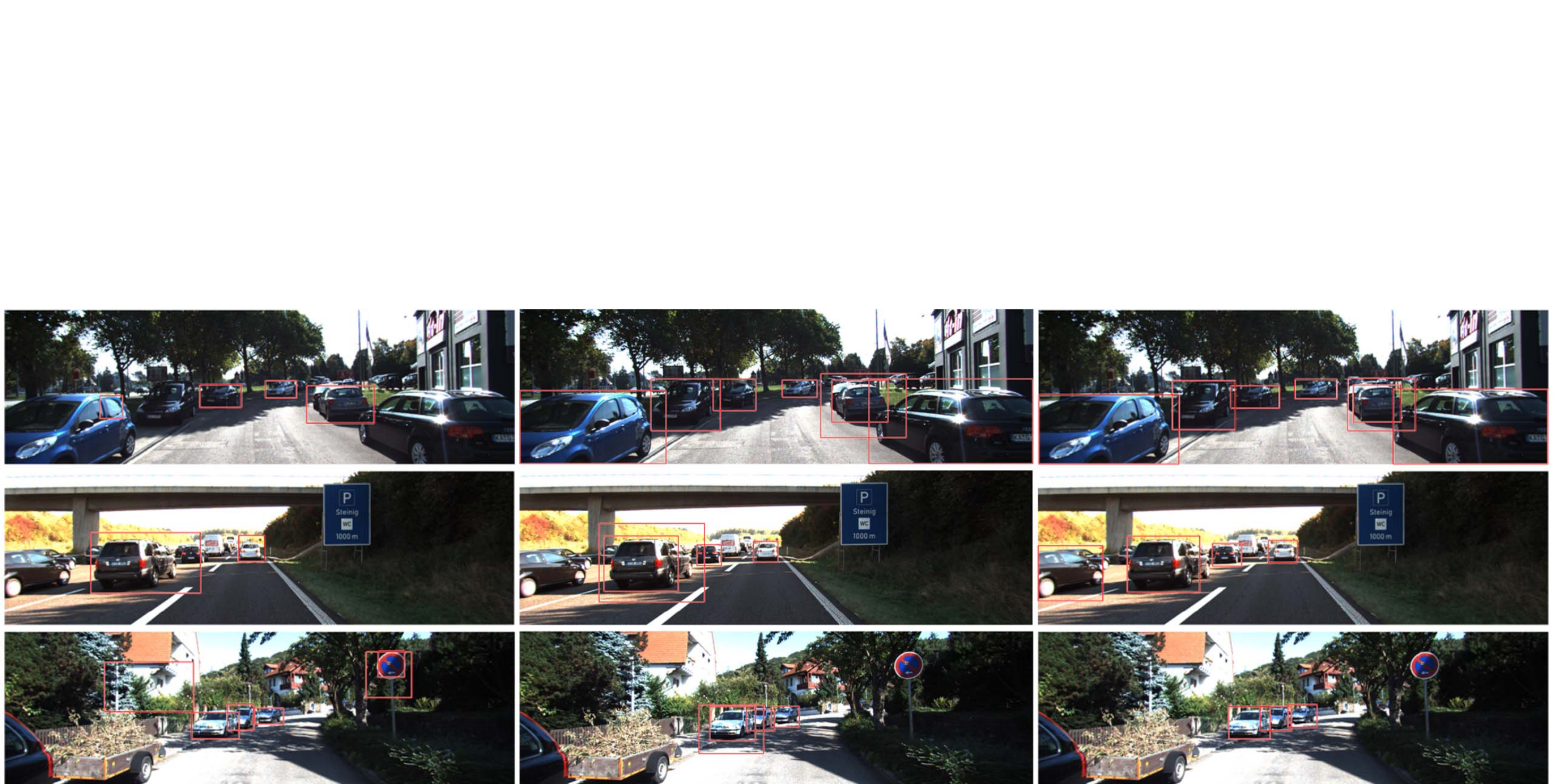}
\caption{Vehicle detections when operating in the FPPI=1 regime. Left: DPM based on the SYNTHIA data considered in this chapter (SRC). Middle: Using the TAR0.1 version of KITTI-Det Train. Right: Adapting SRC by using TAR0.1 when applying SA-SSVM.}
\label{fig:Detections}
\end{figure*}

Figure \ref{fig:Detections} compares vehicle detections based only on the SYNTHIA samples we are using in this chapter, and the result of applying SA-SSVM to them with TAR0.1, in both cases setting the threshold of the model classifier to operate in the FPPI=1 regime. Note how SA-SSVM allows to obtain better results.

TAR0.5 ($X=0.5$; see \fig{plotsX50}) and TAR-ALL basically show the same performance, so $\sim1,582$ images have been annotated without a reward in DPM performance. Of course, although an annotation-training-test loop can be followed to avoid useless vehicle annotations, a priori it is difficult to know when to stop such manual annotations. On the other hand, even using TAR0.5 data, starting with a pre-trained model (SRC) on either SYNTHIA or GTA allows to improve the performance of TAR-ALL, in the case of SYNTHIA by $\sim2$ points with respect to TAR0.5. Looking at SYNTHIA-Sub and GTA, which has a similar number of samples (see \tab{numberSamples}), we can argue that GTA could probably reach the same performance than SYNTHIA if we would have the double of GTA vehicles. In any case, what it is remarkable is that it is more effective to have DPMs pre-trained in virtual worlds than just doubling the number of manually annotated target-domain images, \ie at least assuming a manual annotation procedure free of prior knowledge about the current vehicle detector.

Even for $X=1$ (see \fig{plotsX100}), \ie combining the data used to train TAR-ALL and SRC, pre-training in virtual worlds is able to improve the performance of TAR-ALL alone. SYNTHIA provides us $\sim3$ points of improvement with respect to TAR-ALL, being the overall best result. Using GTA as SRC eventually can provide such improvement too (again, by extrapolation of its performance when comparing to SYNTHIA-Sub).

In summary, the results presented and discussed so far reinforce the take home messages we highlighted in our previous works \cite{MarinCVPR10Learning, VazquezPAMI14Virtual, XuPAMI14Domain}; namely, combining models/data pre-trained in virtual worlds with a reasonable but low amount of real-world data through domain adaptation, is a really practical paradigm worth to explore for learning different kinds of models. As a matter of fact, according to the literature reviewed in sections \ref{sec:needVW-virtualDPM} and \ref{sec:needDA-virtualDPM}, nowadays this approach is being widely adopted by the computer vision community.

Another interesting question that we did not addressed before refers to the degree of photo-realism, \ie if a higher degree would imply to learn more accurate models and eventually not even requiring domain adaptation. This is a very important question since a extreme photo-realism may require hours for rendering a few images, while the degree of photo-realism of the virtual worlds presented here is achieved in real time using a standard modern gamer PC.

In our previous works we already saw that domain adaptation was required even when you train and test with real-world cameras. In other words, domain gap was due to sensor differences (no matter if one of the sensors operates in real or virtual worlds) and the nature of the scenario where train and test images are acquired (typical illumination, background, and pose/view of dynamic objects). Because of this, our believe was that a more photo-realistic world would be just another sensor, still different from real-world, and therefore domain gaps would persists. Note that the experiments presented in this chapter reinforce this hypothesis: (1) using Virtual KITTI and KITTI-Track gives rise to SRC and domain-adapted detectors of similar performance in all the cases, \ie despite the fact that KITTI-Track relies on the same real-world sensor than KITTI-Det Train and Test, while Virtual KITTI consists of synthesized data; (2) moreover, despite the fact that GTA contains images more photo-realistic than SYNTHIA, when using a similar number of samples (SYNTHIA-Sub) we see that the performance of the corresponding SRC and the domain-adapted detectors is basically the same.

Recent works \cite{MovshovitzX16Useful, VeeravasarapuX16Model} reinforce the idea that, once a basic photo-realism is achieved (\ie beyond Lambertian illumination and simplistic object materials), adding more and more photo-realism do not have a relevant impact. Thus, in our opinion from the evidences collected so far, Virtual KITTI and SYNTHIA are sufficiently photo-realistic for the tasks we are addressing (\ie vision-based object detection and image semantic segmentation). 

Another interesting point of analysis is if it is better to just mixing virtual- and real-world data or fine-tuning a pre-trained model on virtual-world data with real-world samples. The former is what we called {\em cool world} \cite{VazquezNIPSDATA11Cool, VazquezPAMI14Virtual}, while SA-SSVM is an example of the later. Because of that we have run a {\em MIX} experiment with SYNTHIA and TAR-ALL, which can be seen in \fig{plotsX100}(c). In this case, we have just mixed the data and run an standard DPM learning procedure. Note that the result is $\sim3$ points worse than using SA-SSVM. Moreover, the training time of MIX is much longer than the one of SA-SSVM, since it uses samples from both domains and training from scratch also requires more iterations to converge. If we extrapolate these experiments to the deep CNNs paradigm, a priori we would think than fine-tuning is the proper approach. However, when working in \cite{RosCVPR16Synthia, RosX16Training}, \ie in semantic segmentation based on deep CNNs, using the appropriate mini-batch scheme to weight the relevance of the samples as a function to their domain (virtual or real), we obtained better performance than with fine-tuning. Therefore, regarding this topic, we have no clear conclusions yet. Of course, the advantage of fine-tuning would be avoiding to revisit the source data; thus, this is a point to keep researching.

Overall, our research and the research presented so far by the computer vision community, led us to insist in the adoption of the paradigm where virtual worlds and domain adaptation techniques are used to train the desired models. Moreover, we think that the degree of photo-realism like the presented already in datasets such as Virtual KITTI and SYNTHIA is sufficient for this task. In addition, although in this chapter we have focused on DPM-based vehicle detection, we think the conclusions can be extrapolated to other computer vision tasks where the visual appearance is important (\eg object detection in general and semantic segmentation). Of course, it is worth to note that at this moment the best results on the KITTI car detection challenge are dominated by approaches based on deep CNNs, providing astonishing high performances in the moderate setting, far beyond DPM approaches. Such benchmark seems to be challenging enough for DPM, but still is a small proportion of the real-world and this will be the real challenge for deep CNNs. Therefore, we also think that our conclusions will be extrapolated from DPM to other powerful models such as deep CNNs when addressing more challenging scenarios; note that in \sect{needDA-virtualDPM} we have mentioned already that even deep CNNs require domain adaptation. On the other hand, what is expected is that deep CNNs would require less domain adaptation than DPM since they are models with more capacity to generalize across domains. 

\section{Conclusion}

In this chapter we have shown how virtual worlds are effective for training visual models when combined with domain adaptation techniques. Although we have focused on DPM and vehicle detection as proof-of-concept, we believe that the conclusions extrapolate to other visual tasks based on more complex models such as deep CNNs. We have presented results which suggest that extreme photo-realism is not necessary, \ie the degree of photo-realism already achieved in datasets such as Virtual KITTI and SYNTHIA is sufficient, provided domain adaptation would be necessary even when relying on more photo-realistic datasets (here GTA). 

Looking into the future, we think a best practice would be to design sets of relatively controlled virtual-world scenarios, designed to train, debug and test visual perception and other AI capabilities (Virtual KITTI and SYNTHIA are examples of this). In other words, with the knowledge accumulated so far, we do not bet for building a gigantic virtual world to try to avoid domain gap issues. This would be really difficult to build and handle. We prefer to pursue domain adaptation to save any existing virtual-to-real world gap. However, we think the research must go into the direction of unsupervised domain adaptation for allowing the systems trained in virtual worlds to self-adapt to real-world scenarios. An example in this line is the approach we presented in \cite{XuICRA16Hierarchical}, where manual annotations were not required to train a domain adapted pedestrian detector for an on-board moving camera setting. However, this approach performs the adaptation off-line, which can be perfectly right for many applications (\eg adapting pre-trained surveillance systems to different places), but the real challenge is to do it on-line. 

\begin{small}
\noindent{\bf Acknowledgments} Authors want to thank the next funding bodies: the Spanish MEC Project TRA2014-57088-C2-1-R, the People Programme (Marie Curie Actions) FP7/2007-2013 REA grant agreement no. 600388, and by the Agency of Competitiveness for Companies of the Government of Catalonia, ACCIO, the Generalitat de Catalunya Project 2014-SGR-1506 and the NVIDIA Corporation for the generous support in the form of different GPU hardware units.
\end{small}

\bibliographystyle{plain}
\bibliography{virtualDPMArxiv}

\begin{thebibliography}{10}

\bibitem{AgarwalACCV06Local}
Ankur Agarwal and Bill Triggs.
\newblock A local basis representation for estimating human pose from cluttered
  images.
\newblock In {\em Asian Conference on Computer Vision ({ACCV})}, 2006.

\bibitem{AubryCVPR14Seeing}
Mathieu Aubry, Daniel Maturana, Alexei Efros, Bryan Russell, and Josef Sivic.
\newblock Seeing {3D} chairs: exemplar part-based 2d-3d alignment using a large
  dataset of {CAD} models.
\newblock In {\em IEEE Conference on Computer Vision and Pattern Recognition
  ({CVPR})}, 2014.

\bibitem{AubryICCV15Understanding}
Mathieu Aubry and {Bryan C.} Russell.
\newblock Understanding deep features with computer-generated imagery.
\newblock In {\em IEEE International Conference on Computer Vision ({ICCV})},
  2015.

\bibitem{BengioPAMI13Representation}
Yoshua Bengio, Aaron Courville, and Pascal Vincent.
\newblock Representation learning: A review and new perspectives.
\newblock {\em Transactions of Pattern Recognition and Machine Analyses
  ({PAMI})}, 35(8):1798--1828, 2013.

\bibitem{BergProcIEEE10All}
{Tamara L.} Berg, Alexander Sorokin, Gang Wang, {David A.} Forsyth, Derek
  Hoiem, Ian Endres, and Ali Farhadi.
\newblock It's all about the data.
\newblock {\em Proceedings of the {IEEE}}, 98(8):1434--1452, 2010.

\bibitem{BochinskiAVSS16Training}
Erik Bochinski, Volker Eiselein, and Thomas Sikora.
\newblock Training a convolutional neural network for multi-class object
  detection using solely virtual world data.
\newblock In {\em Advanced Video and Signal-based Surveillance ({AVSS})}, 2016.

\bibitem{BrostowPRL09Semantic}
{Gabriel J.} Brostow, Julien Fauqueur, and Roberto Cipolla.
\newblock Semantic object classes in video: A high-definition ground truth
  database.
\newblock {\em Pattern Recognition Letters}, 30(20):88--89, 2009.

\bibitem{ButlerECCV12Naturalistic}
{Daniel J.} Butler, Jonas Wulff, {Garrett B.} Stanley, and {Michael J.} Black.
\newblock A naturalistic open source movie for optical flow evaluation.
\newblock In {\em European Conference on Computer Vision ({ECCV})}, 2012.

\bibitem{ChenICCV15DeepDriving}
Chenyi Chen, Ari Seff, {Alain L.} Kornhauser, and Jianxiong Xiao.
\newblock {DeepDriving}: Learning affordance for direct perception in
  autonomous driving.
\newblock In {\em IEEE International Conference on Computer Vision ({ICCV})},
  2015.

\bibitem{ChenCVPR14Beat}
Liang-Chieh Chen, Sanja Fidler, and Raquel Yuille, {Alan L.}~Urtasun.
\newblock Beat the {MTurkers}: Automatic image labeling from weak {3D}
  supervision.
\newblock In {\em IEEE Conference on Computer Vision and Pattern Recognition
  ({CVPR})}, 2014.

\bibitem{ChuECCVTASKCV16Best}
Brian Chu, Vashisht Madhavan, Oscar Beijbom, Judy Hoffman, and Trevor Darrell.
\newblock Best practices for fine-tuning visual classifiers to new domains.
\newblock In {\em European Conference on Computer Vision ({ECCV}), Transferring
  and Adapting Source Knowledge in Computer Vision {(TASK-CV)}}, 2016.

\bibitem{CordtsCVPR16Cityscapes}
Marius Cordts, Mohamed Omran, Sebastian Ramos, Timo Rehfeld, Markus Enzweiler,
  Rodrigo Benenson, Uwe Franke, Stefan Roth, and Bernt Schiele.
\newblock The {Cityscapes} dataset for semantic urban scene understanding.
\newblock In {\em IEEE Conference on Computer Vision and Pattern Recognition
  ({CVPR})}, 2016.

\bibitem{DalalCVPR05Histograms}
Navneet Dalal and Bill Triggs.
\newblock Histograms of oriented gradients for human detection.
\newblock In {\em IEEE Conference on Computer Vision and Pattern Recognition
  ({CVPR})}, pages 886--893, 2005.

\bibitem{DengCVPR09Imagenet}
Jia Deng, Wei Dong, Richard Socher, Li-Jia Li, Kai Li, and Li~Fei-Fei.
\newblock Imagenet: A large-scale hierarchical image database.
\newblock In {\em IEEE Conference on Computer Vision and Pattern Recognition
  ({CVPR})}, 2009.

\bibitem{DollarPAMI12Pedestrian}
Piotr Doll\'ar, Christian Wojek, Bernt Schiele, and Pietro Perona.
\newblock Pedestrian detection: an evaluation of the state of the art.
\newblock {\em Transactions of Pattern Recognition and Machine Analyses
  ({PAMI})}, 34(4):743--761, 2012.

\bibitem{EnzweilerPAMI09Monocular}
Markus Enzweiler and {Dariu M.} Gavrila.
\newblock Monocular pedestrian detection: Survey and experiments.
\newblock {\em Transactions of Pattern Recognition and Machine Analyses
  ({PAMI})}, 31(12):2179--2195, 2009.

\bibitem{FelzenszwalbPAMI10Object}
Pedro~F Felzenszwalb, Ross~B Girshick, David McAllester, and Deva Ramanan.
\newblock Object detection with discriminatively trained part-based models.
\newblock {\em Transactions of Pattern Recognition and Machine Analyses
  ({PAMI})}, 32(9):1627--1645, 2010.

\bibitem{GaidonCVPR16Virtual}
Adrien Gaidon, Qiao Wang, Yohann Cabon, and Eleonora Vig.
\newblock Virtual worlds as proxy for multi-object tracking analysis.
\newblock In {\em IEEE Conference on Computer Vision and Pattern Recognition
  ({CVPR})}, 2016.

\bibitem{GaninICML15Unsupervised}
Yaroslav Ganin and Victor Lempitsky.
\newblock Unsupervised domain adaptation by backpropagation.
\newblock In {\em International Conference on Machine Learning ({ICML})}, pages
  1180--1189, 2015.

\bibitem{GeigerIJRR13Vision}
Andreas Geiger, Philip Lenz, Christoph Stiller, and Raquel Urtasun.
\newblock Vision meets robotics: The {KITTI} dataset.
\newblock {\em International Journal of Robotics Research ({IJRR})},
  32(11):1231--1237, 2016.

\bibitem{GeigerCVPR16Ready}
Andreas Geiger, Philip Lenz, and Raquel Urtasun.
\newblock Are we ready for autonomous driving? the {KITTI} vision benchmark
  suite.
\newblock In {\em IEEE Conference on Computer Vision and Pattern Recognition
  ({CVPR})}, 2012.

\bibitem{GirshickCVPR15Deformable}
Ross Girshick, Forrest Iandola, Trevor Darrell, and Jitendra Malik.
\newblock Deformable part models are convolutional neural networks.
\newblock In {\em IEEE Conference on Computer Vision and Pattern Recognition
  ({CVPR})}, pages 437--446, 2015.

\bibitem{GraumanICCV03Inferring}
Kristen Grauman, Gregory Shakhnarovich, and Trevor Darrell.
\newblock Inferring {3D} structure with a statistical image-based shape model.
\newblock In {\em IEEE International Conference on Computer Vision ({ICCV})},
  2003.

\bibitem{HaeuslerGCPR13Synthesizing}
Ralf Haeusler and Daniel Kondermann.
\newblock Synthesizing real world stereo challenges.
\newblock In {\em German Conference on Pattern Recognition ({GCPR})}, 2013.

\bibitem{HaltakovGCPR13Framework}
Haltakov Haltakov, Christian Unger, and Slobodan Ilic.
\newblock Framework for generation of synthetic ground truth data for driver
  assistance applications.
\newblock In {\em German Conference on Pattern Recognition ({GCPR})}, 2013.

\bibitem{HandaX15SynthCam3D}
Ankur Handa, Viorica Patraucean, Vijay Badrinarayanan, Simon Stent, and Roberto
  Cipolla.
\newblock {SynthCam3D}: Semantic understanding with synthetic indoor scenes.
\newblock {\em CoRR}, arXiv:1505.00171, 2015.

\bibitem{HandaCVPR16Understanding}
Ankur Handa, Viorica Patraucean, Vijay Badrinarayanan, Simon Stent, and Roberto
  Cipolla.
\newblock Understanding real world indoor scenes with synthetic data.
\newblock In {\em IEEE Conference on Computer Vision and Pattern Recognition
  ({CVPR})}, 2016.

\bibitem{HattoriCVPR15Learning}
Hironori Hattori, Vishnu Naresh~Boddeti, Kris~M. Kitani, and Takeo Kanade.
\newblock Learning scene-specific pedestrian detectors without real data.
\newblock In {\em IEEE Conference on Computer Vision and Pattern Recognition
  ({CVPR})}, 2015.

\bibitem{KanevaICCV11Evaluating}
Biliana Kaneva, Antonio Torralba, and {William T.} Freeman.
\newblock Evaluation of image features using a photorealistic virtual world.
\newblock In {\em IEEE International Conference on Computer Vision ({ICCV})},
  2011.

\bibitem{KrizhevskyNIPS12Imagenet}
Alex Krizhevsky, Ilya Sutskever, and Geoff Hinton.
\newblock Image{N}et classification with deep {C}onvolutional {N}eural
  {N}etworks.
\newblock In {\em Annual Conference on Neural Information Processing Systems
  ({NIPS})}, 2012.

\bibitem{LaiRSS09Laser}
Kevin Lai and Dieter Fox.
\newblock {3D} laser scan classification using web data and domain adaptation.
\newblock In {\em Robotics: Science and Systems}, 2009.

\bibitem{LaiIJRR10Object}
Kevin Lai and Dieter Fox.
\newblock Object recognition in {3D} point clouds using web data and domain
  adaptation.
\newblock {\em International Journal of Robotics Research ({IJRR})},
  29(8):1019--1037, 2010.

\bibitem{ChenECCV12Recognizing}
Wenbin Li and Mario Fritz.
\newblock Recognizing materials from virtual examples.
\newblock In {\em European Conference on Computer Vision ({ECCV})}, 2012.

\bibitem{LinECCV14Microsoft}
Tsung-Yi Lin, Michael Maire, Serge Belongie, James Hays, Pietro Perona, Deva
  Ramanan, Piotr Doll\'{a}r, and C.Lawrence Zitnick.
\newblock Microsoft {COCO}: Common objects in context.
\newblock In {\em European Conference on Computer Vision ({ECCV})}, pages
  740--755, 2014.

\bibitem{LlarguesESA14Artificial}
{Joan M.} Llargues, Juan Peralta, Raul Arrabales, Manuel Gonz\'{a}lez, Paulo
  Cortez, and {Antonio M.} L\'opez.
\newblock Artificial intelligence approaches for the generation and assessment
  of believable human-like behaviour in virtual characters.
\newblock {\em Expert Systems With Applications}, 41(16):7281--7290, 2014.

\bibitem{MarinCVPR10Learning}
Javier Mar{\'i}n, David V\'azquez, David Ger\'onimo, and {Antonio M.} L\'opez.
\newblock Learning appearance in virtual scenarios for pedestrian detection.
\newblock In {\em IEEE Conference on Computer Vision and Pattern Recognition
  ({CVPR})}, 2010.

\bibitem{MassaCVPR16Deep}
Francisco Massa, {Bryan C.} Russell, and Mathieu Aubry.
\newblock Deep exemplar {2D-3D} detection by adapting from real to rendered
  views.
\newblock In {\em IEEE Conference on Computer Vision and Pattern Recognition
  ({CVPR})}, 2016.

\bibitem{MayerCVPR16Large}
Nikolaus Mayer, Eddy Ilg, Philip Hausser, Philipp Fischer, Daniel Cremers,
  Alexey Dosovitskiy, and Thomas Brox.
\newblock A large dataset to train convolutional networks for disparity,
  optical flow, and scene flow estimation.
\newblock In {\em IEEE Conference on Computer Vision and Pattern Recognition
  ({CVPR})}, 2016.

\bibitem{MeisterCEMT11Real}
Stephan Meister and Daniel Kondermann.
\newblock Real versus realistically rendered scenes for optical flow
  evaluation.
\newblock In {\em Conference on Electronic Media Technology ({CEMT})}, 2011.

\bibitem{MnihNIPSWDL13Playing}
Volodymyr Mnih, Koray Kavukcuoglu, David Silver, Alex Graves, Ioannis
  Antonoglou, Daan Wierstra, and Martin Riedmiller.
\newblock Playing {Atari} with deep reinforcement learning.
\newblock In {\em Annual Conference on Neural Information Processing Systems
  ({NIPS}), Workshop on Deep Learning}, 2013.

\bibitem{MovshovitzX16Useful}
Yair {Movshovitz-Attias}, Takeo Kanade, and Yaser Sheikh.
\newblock How useful is photo-realistic rendering for visual learning?
\newblock {\em CoRR}, arXiv:1603.08152, 2016.

\bibitem{OnkarappaMTA15Synthetic}
Naveen Onkarappa and {Angel D.} Sappa.
\newblock Synthetic sequences and ground-truth flow field generation for
  algorithm validation.
\newblock {\em Multimedia Tools and Applications}, 74(9):3121--3135, 2015.

\bibitem{PanaredaBustoBMVC15Adaptation}
Pau Panareda-Busto, Joerg Liebelt, and Juergen Gall.
\newblock Adaptation of synthetic data for coarse-to-fine viewpoint refinement.
\newblock In {\em BMVA British Machine Vision Conference ({BMVC})}, 2015.

\bibitem{PaponICCV15Semantic}
Jeremie Papon and Markus Schoeler.
\newblock Semantic pose using deep networks trained on synthetic {RGB-D}.
\newblock In {\em IEEE International Conference on Computer Vision ({ICCV})},
  2015.

\bibitem{PengICCV15Learning}
Xingchao Peng, Baochen Sun, Karim Ali, and Kate Saenko.
\newblock Learning deep object detectors from {3D} models.
\newblock In {\em IEEE International Conference on Computer Vision ({ICCV})},
  2015.

\bibitem{PepikCVPR12Teaching}
Bojan Pepik, Michael Stark, Peter Gehler, and Bernt Schiele.
\newblock Teaching {3D} geometry to deformable part models.
\newblock In {\em IEEE Conference on Computer Vision and Pattern Recognition
  ({CVPR})}, 2012.

\bibitem{PishchulinCVPR11Learning}
Leonid Pishchulin, Arjun Jain, Christian Wojek, Mykhaylo Andriluka, Thorsten
  Thorm\"ahlen, and Bernt Schiele.
\newblock Learning people detection models from few training samples.
\newblock In {\em IEEE Conference on Computer Vision and Pattern Recognition
  ({CVPR})}, 2011.

\bibitem{AmazonMT}
\protect{Amazon Mechanical Turk}.
\newblock \texttt{http://www.mturk.com}.

\bibitem{RichterECCV16Playing}
{Stephan R.} Richter, Vibhav Vineet, Stefan Roth, and Koltun Vladlen.
\newblock Playing for data: Ground truth from computer games.
\newblock In {\em European Conference on Computer Vision ({ECCV})}, 2016.

\bibitem{RosCVPR16Synthia}
German Ros, Laura Sellart, Joanna Materzyska, David V{\'a}zquez, and {Antonio
  M.} L{\'o}pez.
\newblock The {SYNTHIA} dataset: a large collection of synthetic images for
  semantic segmentation of urban scenes.
\newblock In {\em IEEE Conference on Computer Vision and Pattern Recognition
  ({CVPR})}, 2016.

\bibitem{RosX16Training}
German Ros, Simon Stent, {Pablo F.} Alcantarilla, and Tomoki Watanabe.
\newblock Training constrained deconvolutional networks for road scene semantic
  segmentation.
\newblock {\em CoRR}, arXiv:1603.08152, 2016.

\bibitem{RozantsevCVIU15Rendering}
Artem Rozantsev, Vincent Lepetit, and Pascal Fua.
\newblock On rendering synthetic images for training an object detector.
\newblock {\em Computer Vision and Image Understanding}, 137:24--37, 2015.

\bibitem{RussellIJCV08LabelMe}
{Bryan C.} Russell, Antonio Torralba, {Kevin P.} Murphy, and {William T.}
  Freeman.
\newblock {LabelMe}: a database and web-based tootl for image annotation.
\newblock {\em International Journal of Computer Vision ({IJCV})},
  77(1--3):157--173, 2008.

\bibitem{SanchezIJCV13Image}
Jorge S\'{a}nchez, Florent Perronnin, Thomas Mensink, and Jakob Verbeek.
\newblock Image classification with the fisher vector: Theory and practice.
\newblock {\em International Journal of Computer Vision ({IJCV})},
  105(3):222--245, 2013.

\bibitem{SatkinBMVC10Back}
Scott Satkin, Michael Goesele, and Bernt Schiele.
\newblock Back to the future: Learning shape models from {3D CAD} data.
\newblock In {\em BMVA British Machine Vision Conference ({BMVC})}, 2010.

\bibitem{SatkinBMVC12Data}
Scott Satkin, Jason Lin, and Martial Hebert.
\newblock Data-driven scene understanding from {3D} models.
\newblock In {\em BMVA British Machine Vision Conference ({BMVC})}, 2012.

\bibitem{SchelsICMR11Synthetically}
Johannes Schels, J\"{o}rg Liebelt, Klaus Schertler, and Rainer Lienhart.
\newblock Synthetically trained multi-view object class and viewpoint detection
  for advanced image retrieval.
\newblock In {\em International Conference on Multimedia Retrieval ({ICMR})},
  2011.

\bibitem{ShafaeiBMVC16Play}
Alireza Shafaei, {James J.} Little, and Mark Schmidt.
\newblock Play and learn: Using video games to train computer vision models.
\newblock In {\em BMVA British Machine Vision Conference ({BMVC})}, 2016.

\bibitem{ShottonCVPR11Realtime}
Jamie Shotton, Andrew Fitzgibbon, Mat Cook, Toby Sharp, Mark Finocchio, Richard
  Moore, Alex Kipmanand, and Andrew Blake.
\newblock Real-time human pose recognition in parts from a single depth image.
\newblock In {\em IEEE Conference on Computer Vision and Pattern Recognition
  ({CVPR})}, 2011.

\bibitem{SocarrasICCVVisDA13Adapting}
Yainuvis Socarras, Sebastian Ramos, David V\'azquez, {Antonio M.} L\'opez, and
  Theo Gevers.
\newblock Adapting pedestrian detection from synthetic to far infrared images.
\newblock In {\em IEEE International Conference on Computer Vision ({ICCV}),
  Visual Domain Adaptation and Dataset Bias {(ICCV-VisDA)}}, 2013.

\bibitem{SuICCV16Render}
Hao Su, {Charles R.} Qi, Yangyan Yi, and Leonidas Guibas.
\newblock Render for {CNN}: viewpoint estimation in images using {CNNs} trained
  with rendered {3D} model views.
\newblock In {\em IEEE International Conference on Computer Vision ({ICCV})},
  2015.

\bibitem{SuICCV163D}
Hao Su, Fan Wang, Yangyan Yi, and Leonidas Guibas.
\newblock {3D}-assisted feature synthesis for novel views of an object.
\newblock In {\em IEEE International Conference on Computer Vision ({ICCV})},
  2015.

\bibitem{SunBMVC14Virtual}
Baochen Sun and Kate Saenko.
\newblock From virtual to reality: Fast adaptation of virtual object detectors
  to real domains.
\newblock In {\em BMVA British Machine Vision Conference ({BMVC})}, 2014.

\bibitem{TaylorCVPR07OVVV}
{Geoffrey R.} Taylor, {Andrew J.} Chosak, and {Paul C.} Brewer.
\newblock {OVVV}: Using virtual worlds to design and evaluate surveillance
  systems.
\newblock In {\em IEEE Conference on Computer Vision and Pattern Recognition
  ({CVPR})}, 2007.

\bibitem{TommasiGCPR15Deeper}
Tatiana Tommasi, Novi Patricia, Barbara Caputo, and Tinne Tuytelaars.
\newblock A deeper look at dataset bias.
\newblock In {\em German Conference on Pattern Recognition ({GCPR})}, 2015.

\bibitem{TorralbaCVPR11Unbiased}
Antonio Torralba and Alexei~A. Efros.
\newblock Unbiased look at dataset bias.
\newblock In {\em IEEE Conference on Computer Vision and Pattern Recognition
  ({CVPR})}, 2011.

\bibitem{TzengICCV15Simultaneous}
Eric Tzeng, Judy Hoffman, Trevor Darrell, and Kate Saenko.
\newblock Simultaneous deep transfer across domains and tasks.
\newblock In {\em IEEE International Conference on Computer Vision ({ICCV})},
  pages 4068--4076, 2015.

\bibitem{VazquezPAMI14Virtual}
David Vazquez, Antonio~M. L\'{o}pez, Javier Mar\'{i}n, Daniel Ponsa, and David
  Ger\'{o}nimo.
\newblock Virtual and real world adaptation for pedestrian detection.
\newblock {\em Transactions of Pattern Recognition and Machine Analyses
  ({PAMI})}, 36(4):797 -- 809, 2014.

\bibitem{VazquezMIIVAIS13Interactive}
David V\'azquez, {Antonio M.} L\'opez, Daniel Ponsa, and David Ger\'onimo.
\newblock Interactive training of human detectors.
\newblock In {Angel D.} Sappa and Jordi Vitri\`a, editors, {\em Multimodal
  Interaction in Image and Video Applications Intelligent Systems}, pages
  169--184. Springer, 2013.

\bibitem{VazquezNIPSDATA11Cool}
David V\'azquez, {Antonio M.} L\'opez, Daniel Ponsa, and Javier Mar\'{\i}n.
\newblock Cool world: domain adaptation of virtual and real worlds for human
  detection using active learning.
\newblock In {\em Annual Conference on Neural Information Processing Systems
  ({NIPS}), Workshop on Domain Adaptation: Theory and Applications}, 2011.

\bibitem{VazquezICMI11Virtual}
David V\'azquez, {Antonio M.} L\'opez, Daniel Ponsa, and Javier Mar\'{\i}n.
\newblock Virtual worlds and active learning for human detection.
\newblock In {\em International Conference on Multimodal Interaction ({ICMI})},
  2011.

\bibitem{VedantamICCV15Learning}
Ramakrishna Vedantam, Xiao Lin, Tanmay Batra, {C. Lawrence} Zitnick, and Devi
  Parikh.
\newblock Learning common sense through visual abstraction.
\newblock In {\em IEEE International Conference on Computer Vision ({ICCV})},
  2015.

\bibitem{VeeravasarapuX15Model}
{V.S.R.} Veeravasarapu, {Rudra Narayan} Hota, Constantin Rothkopf, and Ramesh
  Visvanathan.
\newblock Model validation for vision systems via graphics simulation.
\newblock {\em CoRR}, arXiv:1512.01401, 2015.

\bibitem{VeeravasarapuX15Simulations}
{V.S.R.} Veeravasarapu, {Rudra Narayan} Hota, Constantin Rothkopf, and Ramesh
  Visvanathan.
\newblock Simulations for validation of vision systems.
\newblock {\em CoRR}, arXiv:1512.01030, 2015.

\bibitem{VeeravasarapuX16Model}
{V.S.R.} Veeravasarapu, Constantin Rothkopf, and Ramesh Visvanathan.
\newblock Model-driven simulations for deep convolutional neural networks.
\newblock {\em CoRR}, arXiv:1605.09582, 2016.

\bibitem{ViolaCVPR01Rapid}
Paul Viola and Michael Jones.
\newblock Rapid object detection using a boosted cascade of simple features.
\newblock In {\em IEEE Conference on Computer Vision and Pattern Recognition
  ({CVPR})}, 2001.

\bibitem{XuPAMI14Domain}
Jiaolong Xu, Sebastian Ramos, David V{\'a}zquez, and {Antonio M.} L{\'o}pez.
\newblock Domain adaptation of deformable part-based models.
\newblock {\em Transactions of Pattern Recognition and Machine Analyses
  ({PAMI})}, 36(12):2367--2380, 2014.

\bibitem{XuIJCV16Hierarchical}
Jiaolong Xu, Sebastian Ramos, David V{\'a}zquez, and {Antonio M.} L{\'o}pez.
\newblock Hierarchical adaptive structural {SVM} for domain adaptation.
\newblock {\em International Journal of Computer Vision ({IJCV})},
  119(2):159--178, 2016.

\bibitem{XuITS14Learning}
Jiaolong Xu, David V{\'a}zquez, {Antonio M.} L{\'o}pez, Javier Mar{\'i}n, and
  Daniel Ponsa.
\newblock Learning a part-based pedestrian detector in a virtual world.
\newblock {\em Transactions on Intelligent Transportation Systems ({ITS})},
  15(5):2121--2131, 2014.

\bibitem{XuICRA16Hierarchical}
Jiaolong Xu, David V{\'a}zquez, Krystian Mikolajczyk, and {Antonio M.}
  L{\'o}pez.
\newblock Hierarchical online domain adaptation of deformable part-based
  models.
\newblock In {\em International Conference on Robotics and Automation
  ({ICRA})}, 2016.

\bibitem{ZhuIJCV16Need}
Xiangxin Zhu, Carl Vondrick, Charless~C. Fowlkes, and Deva Ramanan.
\newblock Do we need more training data?
\newblock {\em International Journal of Computer Vision ({IJCV})},
  119(1):76–--92, 2016.

\bibitem{ZhuX16Target}
Yuke Zhu, Roozbeh Mottaghi, Eric Kolve, {Joseph J.} Lim, and Abhinav Gupta.
\newblock Target-driven visual navigation in indoor scenes using deep
  reinforcement learning.
\newblock {\em CoRR}, arXiv:1609.05143, 2016.

\bibitem{ZitnickPAMI16Adopting}
{C. Lawrence} Zitnick, Ramakrishna Vedantam, and Devi Parikh.
\newblock Adopting abstract images for semantic scene understanding.
\newblock {\em Transactions of Pattern Recognition and Machine Analyses
  ({PAMI})}, 38(4):627--638, 2016.

\end{thebibliography}

\end{document}